%% file: main.tex
\documentclass[10pt,twocolumn,letterpaper]{article}

\usepackage[pagenumbers]{cvpr} % To force page numbers, e.g. for an arXiv version

\input{preamble}

\definecolor{cvprblue}{rgb}{0.21,0.49,0.74}
\usepackage[pagebackref,breaklinks,colorlinks,allcolors=cvprblue]{hyperref}

\def\paperID{*****} % *** Enter the Paper ID here
\def\confName{CVPR}
\def\confYear{2026}

\title{Uncertainty-Aware World Model for Aerial Image-Goal Navigation}

\author{Deyi Zhu\textsuperscript{*}, 
Haoyu Fan\textsuperscript{*}, 
Yinan Zhu, 
Weichen Zhang, 
Shilin Ma, 
Xinlei Chen, 
Yansong Tang\textsuperscript{\textdagger}\\
Tsinghua Shenzhen International Graduate School, Tsinghua University\\
{\tt\small zhudy21@mails.tsinghua.edu.cn, tang.yansong@sz.tsinghua.edu.cn}\\
\small Project Page: \tt\textcolor{cvprblue}{https://duryi.github.io/UA-NWM-Project-Page}
}

\begin{document}
\maketitle
{
\renewcommand{\thefootnote}{\fnsymbol{footnote}}
\footnotetext[1]{Equal contribution.}
\footnotetext[2]{Corresponding author.}
}
\input{sec/0_abstract}
\input{sec/1_main}
\clearpage
{
    \small
    \bibliographystyle{ieeenat_fullname}
    \bibliography{main}
}

% WARNING: do not forget to delete the supplementary pages from your submission 
\input{sec/X_suppl}

\end{document}

%% file: preamble.tex
\usepackage{booktabs}
\usepackage{multirow}
\usepackage{array}
\usepackage{amssymb}

\usepackage[table]{xcolor}
\usepackage{tabularx}

%% file: sec/0_abstract.tex
\begin{abstract}
Aerial image-goal navigation requires an unmanned aerial vehicle (UAV) to reach a target location specified by a goal image. Existing world-model-based methods rank candidate trajectories using predicted futures, but typically rely on only one or a few point predictions, which is inadequate for large-scale outdoor environments with substantial future-state uncertainty. To address this limitation, we propose the \textbf{U}ncertainty-\textbf{A}ware \textbf{N}avigation \textbf{W}orld \textbf{M}odel (\textbf{UA-NWM}), an efficient latent world model for aerial image-goal navigation, which formulates trajectory scoring as conditional out-of-distribution detection. UA-NWM represents plausible futures with an uncertainty subspace and decomposes the prediction--goal discrepancy into uncertainty-explainable and unexplainable components. Only the unexplainable residual is used for scoring, enabling robust selection without multiple future samples. Extensive experiments demonstrate that UA-NWM consistently outperforms existing navigation world models while maintaining low inference latency. Real-world UAV experiments further validate its practical applicability.
\end{abstract}

%% file: sec/1_main.tex
\section{Introduction}

Image-goal navigation maps current observation and a goal image to actions that guide an embodied agent toward the target position. Conventional methods~\cite{nomad,vint,gnm,flownav,navibridger} directly predict actions or trajectories, whereas recent approaches~\cite{nwm,anwm,mwm,raenwm,relnwm,onestepwm} use world models to score candidate trajectories by predicting their future states, and select the one most likely to reach the goal for execution.

Existing world models can be broadly categorized as stochastic or deterministic. Stochastic world models~\cite{genie, astra, gwm, cosmos3} can generate different plausible futures, but are typically computationally expensive. Some studies~\cite{dinowm, dinoworld, dinoforesight, lawam} therefore perform prediction in the latent spaces of vision foundation models (VFMs)~\cite{dinov2, dinov3} to improve efficiency, but most of them remain deterministic, and can only produce a single estimate, rather than representing the full future-state distribution. Recent methods~\cite{vfmf, deltatok, svg} begin exploring stochastic prediction in VFM latent spaces.

\begin{figure}[t]
\centering
\includegraphics[width=\linewidth]{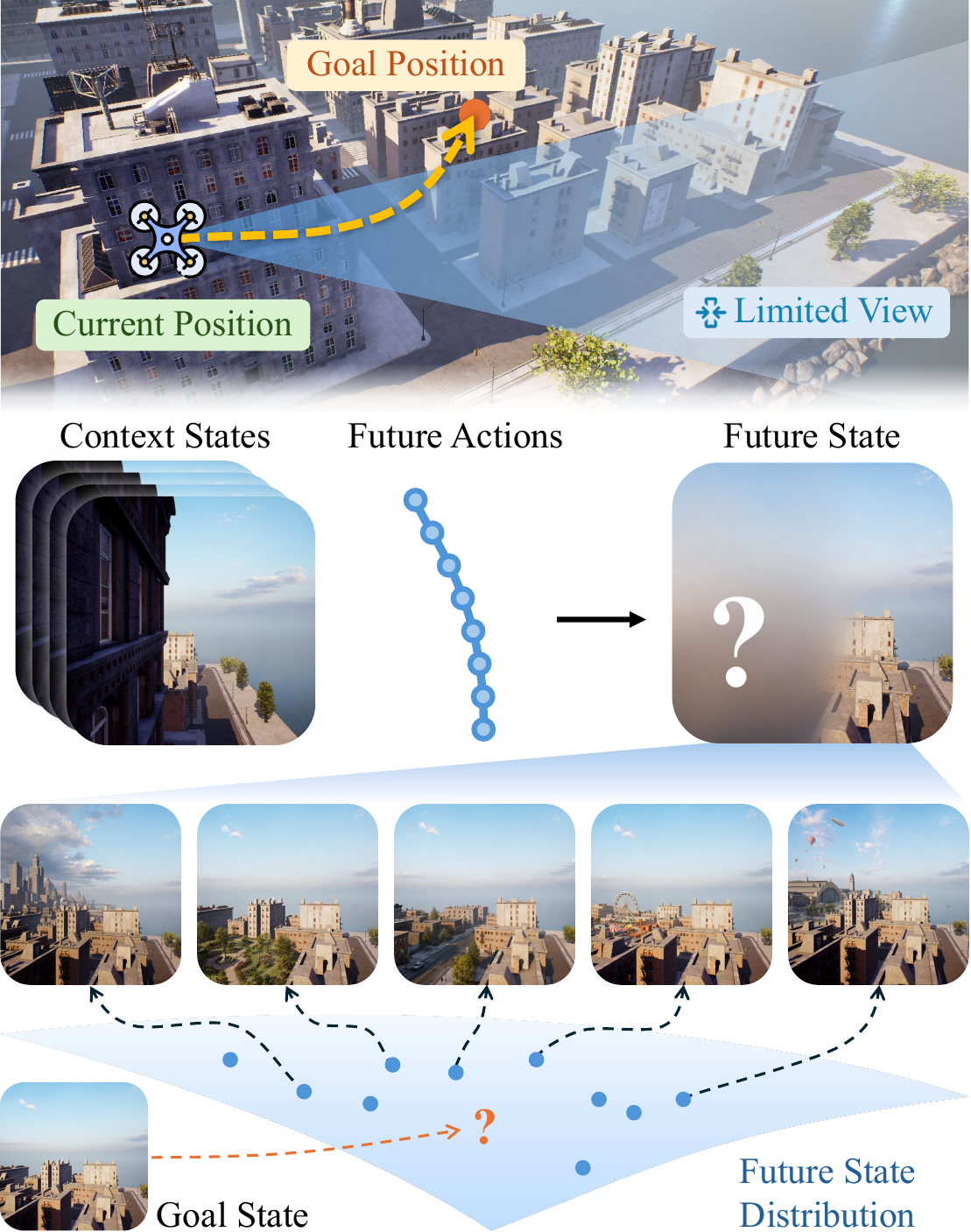}
\caption{Illustration of \textit{future-state uncertainty}.}
\label{fig_teaser}
\end{figure}

\begin{figure*}[t]
\centering
\includegraphics[width=\textwidth]{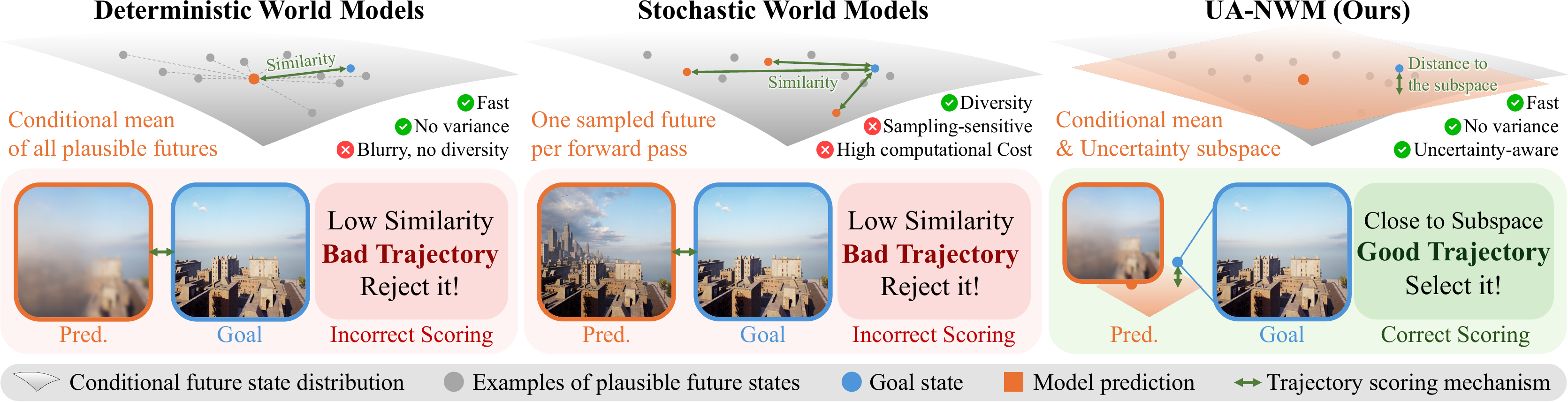}
\caption{Comparison of UA-NWM and previous navigation world models in terms of future prediction and trajectory scoring.}
\label{fig_compare}
\end{figure*}

Despite these advances, a critical limitation remains---\textit{future-state uncertainty} is not fully exploited in world-model-based trajectory scoring for navigation. This issue is especially pronounced in large-scale outdoor scenes, such as urban aerial navigation, where long-horizon motion and unseen regions yield multiple plausible futures under the same context and actions. As illustrated in Figure~\ref{fig_teaser}, occlusion by the building leaves the future appearance of the left region ambiguous. Under the manifold hypothesis~\cite{manifoldhypothesis}, these plausible futures concentrate around a low-dimensional manifold. Trajectory scoring can therefore be viewed as determining whether the goal image lies within the conditional future-state distribution.

However, existing navigation world models typically compare the goal with only one or a few point predictions, as shown in Figure~\ref{fig_compare}. Deterministic models~\cite{lsnwm,relnwm} collapse multiple plausible futures into a single, often over-smoothed estimate that may even lie outside the high-density region of the true distribution, making it unreliable for scoring. Stochastic models~\cite{nwm,raenwm,mwm,onestepwm} can capture diverse plausible futures, but determining whether the goal belongs to this distribution requires sufficient samples to approximate it, incurring high computational cost. Relying on only one or a few samples makes scoring sensitive to sampling randomness. Consequently, under \textit{future-state uncertainty}, both paradigms are prone to assigning low scores to promising candidates and incorrectly rejecting them, thereby degrading navigation performance.

To address these limitations, we formulate trajectory scoring as a conditional out-of-distribution (OOD) detection problem. Each candidate trajectory defines a conditional future-state distribution, where the goal state is in-distribution if the trajectory can reach it and out-of-distribution otherwise. Since this distribution can only be modeled implicitly, directly determining whether the goal lies within it remains challenging. Instead of sampling multiple futures, we approximate this distribution with a predicted uncertainty subspace, as shown in Figure~\ref{fig_compare}. Trajectories are scored by the distance between the goal state and this subspace, avoiding reliance on either an over-smoothed deterministic prediction or an arbitrary stochastic sample.

Based on this formulation, we propose the \textbf{U}ncertainty-\textbf{A}ware \textbf{N}avigation \textbf{W}orld \textbf{M}odel (\textbf{UA-NWM}) for aerial image-goal navigation. UA-NWM uses a lightweight deterministic backbone to predict future representations in the DINOv3~\cite{dinov3} latent space, and introduces a Hierarchical Error Projection (HEP) module to model an uncertainty subspace around each prediction. HEP decomposes the discrepancy between the predicted and goal representations into parallel and orthogonal components, corresponding to uncertainty-explainable variation and unexplained residual error, respectively. Only the orthogonal residual is used for trajectory scoring, avoiding penalties on plausible uncertainty-induced deviations.

We construct a high-quality dataset from existing aerial navigation benchmarks and train UA-NWM with a staged optimization scheme. Extensive offline and online experiments demonstrate that UA-NWM substantially outperforms existing methods while maintaining low inference latency, with the uncertainty-aware HEP consistently improving trajectory-scoring accuracy. Real-world UAV experiments further validate its effectiveness for real application.

Our main contributions are summarized as follows:
\begin{itemize}
\item We propose UA-NWM, an uncertainty-aware navigation world model that formulates trajectory scoring as conditional OOD detection and represents plausible future states with an uncertainty subspace.
\item We introduce a Hierarchical Error Projection (HEP) module that decomposes the prediction--goal discrepancy into uncertainty-explainable and unexplainable components, using only the latter for robust trajectory scoring.
\item UA-NWM achieves state-of-the-art performance in both offline and online evaluations while maintaining low inference latency. Its effectiveness and practicality are further demonstrated through real-world UAV experiments.
\end{itemize}

\section{Related Work}

\subsection{Image-Goal Navigation}
Image-goal navigation guides an agent to an image-specified position, supporting applications such as delivery, inspection, and rescue when precise coordinates are unavailable.

Classical methods rely on mapping or localization for collision-free planning~\cite{bonin2008visual,zhang2022survey, complexmodularlearning,iglnav}, but may accumulate errors in complex scenes. Learning-based approaches directly predict actions or waypoints from observations and goals. Early methods~\cite{zhu2017target} use reinforcement learning, while recent studies~\cite{gnm,vint,unigoal} improve generalization with heterogeneous robot data, Transformer policies, or LLM backbones. Generative policies~\cite{nomad,flownav,navidiffusor,navibridger,geninav} further model multimodal action or trajectory distributions.

Recent work has extended image-goal navigation to UAVs, but they mainly target indoor exploration~\cite{sign, flyingimage}, language guidance~\cite{vlfly}, or local pose alignment~\cite{pairuav}. These methods generally lack action-conditioned prediction of multi-step visual outcomes for candidate trajectories, limiting reliable long-horizon planning in complex outdoor environments.

\subsection{World Models for Navigation}

World models have been widely adopted in embodied tasks~\cite{manigaussian,manigaussian++,clap}, while the broader ideas of future prediction and prediction-guided selection have also been explored across diverse visual applications~\cite{safepruner,samosa,taihri,vgrefiner,sam2love,iterprime}. For navigation world models allow agents to score candidate actions by predicting their future visual consequences. Early work~\cite{pathdreamer,dreamwalker,dreamnav} explored future-scene imagination for indoor and long-horizon navigation, while NWM~\cite{nwm} introduced action-conditioned video prediction for trajectory scoring. Following NWM, recent methods have improved prediction efficiency and representation quality. One-Step WM~\cite{onestepwm} generates future observations in a single step, while LS-NWM~\cite{lsnwm}, ReL-NWM~\cite{relnwm}, and RAE-NWM~\cite{raenwm} perform prediction in dense VFM feature spaces. Another line of work~\cite{uniwm,navwm,navwam} jointly integrates future prediction with trajectory planning. In aerial navigation, WorldVLN~\cite{worldvln} and ImagineUAV~\cite{imagineuav} focus on language-conditioned flight, while ANWM~\cite{anwm} uses depth-assisted future-frame projection for image-goal navigation.

Despite these advances, existing methods typically score each trajectory using only one or a few predicted futures. Such point-wise comparison cannot distinguish plausible future variations from action-inconsistent errors. Deterministic models collapse multiple outcomes into a single blurry estimate, while stochastic models remain sensitive to sampling variance. This motivates an explicit representation of the future-state distribution for trajectory scoring.

\section{Preliminaries}

In image-goal navigation, a world model can serve as a trajectory ranker rather than directly predicting the next control command. At time $t$, the agent observes a context of $C$ images,
$O_{t-C+1:t}=\{o_{t-C+1},\ldots,o_t\}$, and scores action sequences over a horizon of $H$ steps. Each candidate sequence is denoted by
$A=\{a_t,\ldots,a_{t+H-1}\}$. A proposal policy or sampling-based planner generates a finite candidate set
$\mathcal{A}=\{A^{(1)},\ldots,A^{(Q)}\}$. For each candidate trajectory, the world model predicts its terminal visual representation and evaluates its compatibility with the goal image $o_g$.
Formally, the world model assigns each candidate a trajectory cost
\begin{equation}
s(A,o_g \mid O_{t-C+1:t}) =
D\left(
F_\theta(O_{t-C+1:t},A),
\phi(o_g)
\right),
\end{equation}
where $\phi(\cdot)$ is a frozen visual feature extractor, $F_\theta$ predicts the future feature representation conditioned on the observation context and candidate actions, and $D(\cdot,\cdot)$ measures their discrepancy. The optimal action sequence is then selected as
\begin{equation}
    A^\star = \mathrm{argmin}_{A \in \mathcal{A}} s(A,o_g \mid O_{t-C+1:t}).
\end{equation}
Existing world-model rankers typically define $D$ as a point-to-point distance. Pixel-space models compare the generated future image with the goal using perceptual metrics such as LPIPS~\cite{lpips}, whereas latent-space models measure the cosine or Euclidean distance between feature representations.
This formulation also applies to standalone planning, where $\mathcal{A}$ is iteratively refined by a sampling-based optimizer such as the cross-entropy method (CEM)~\cite{cem}, with the cost $s$ serving as the optimization objective.

\section{Method}

\begin{figure*}[t]
\centering
\includegraphics[width=\textwidth]{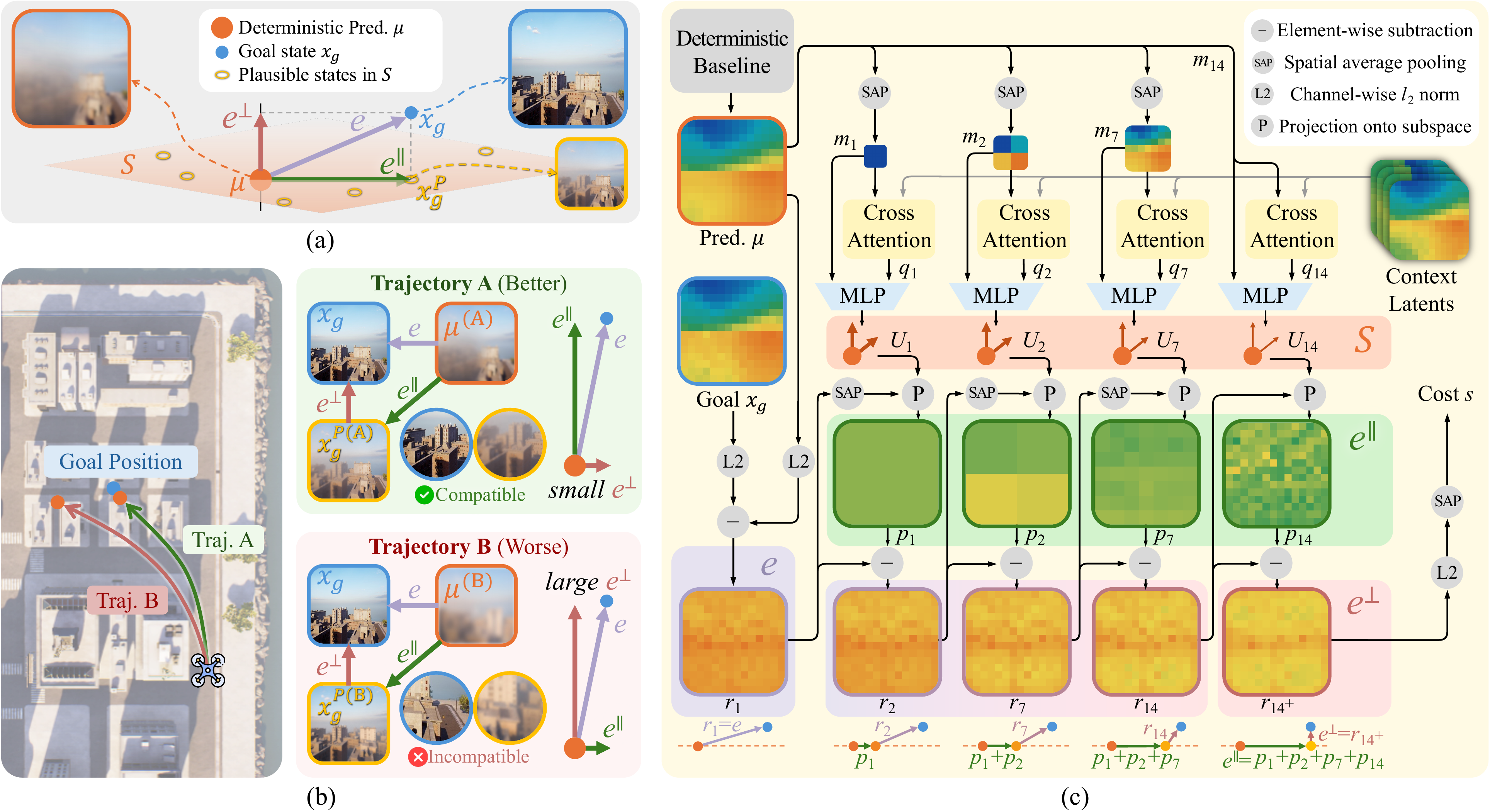}
\caption{(a) Illustration of the uncertainty-aware trajectory scoring idea. Conventional deterministic methods score based on $e$. We divide $e$ as $e^{\perp}$ and $e^{\parallel}$, and use only $e^{\perp}$ for scoring. (b) An example of how uncertainty-aware trajectory scoring can improve navigation performance. (c) The overall pipeline of our proposed Hierarchical Error Projection (HEP) module.}
\label{fig_hyperplane}
\end{figure*}

UA-NWM scores candidate trajectories by measuring goal compatibility with the future-state distribution conditioned on the current context and each candidate action sequence. We first formulate the idea of uncertainty-aware trajectory scoring and introduce the Hierarchical Error Projection (HEP) module to implement it, then describe the deterministic latent world model backbone and training procedure.

\subsection{Uncertainty-Aware Trajectory Scoring}

Given context observations $O_{t-C+1:t}$ and a candidate action sequence $A_{t:t+H-1}$, the latent world model makes a deterministic prediction of the final-horizon DINO feature map:
\begin{equation}
    \mu = F_\theta(O_{t-C+1:t}, A_{t:t+H-1}),
\end{equation}
where $\mu$ contains $N$ patch tokens. Let $x_g=\phi(o_g)$ denote the DINO feature map of the goal $o_g$. Their normalized discrepancy is
\begin{equation}
    e = \mathrm{norm}(x_g) - \mathrm{norm}(\mu),
\end{equation}
where $\mathrm{norm}(\cdot)$ denotes channel-wise $\ell_2$ normalization. Previous deterministic scorers penalize the full discrepancy $e$.

We instead formulate trajectory scoring as conditional OOD detection. For each candidate trajectory, its future-state distribution defines an in-distribution region in the DINO feature space. As shown in Figure~\ref{fig_hyperplane}(a), we approximate this region with an uncertainty subspace $\mathcal{S}$ around $\mu$ and decompose $e$ into an explainable component $e^\parallel$ and an orthogonal residual $e^\perp$. Only the unexplained residual $e^\perp$ is penalized:
\begin{equation}
    s(A_{t:t+H-1}, o_g \mid O_{t-C+1:t})
    =
    \frac{1}{N}\sum_{i=1}^{N}
    \left\| e_i^\perp \right\|_2^2,
\end{equation}
where $e_i^\perp$ is the residual at the $i$-th patch of $e^\perp$. 

Therefore, a candidate receives a low cost when the goal differs from the predicted mean primarily along plausible uncertainty directions, and a high cost when the discrepancy cannot be explained by $\mathcal{S}$. In Figure~\ref{fig_hyperplane}(b), trajectories A and B have similar total discrepancies $e$. For trajectory A, most of the discrepancy between $\mu^{(A)}$ and $x_g$ can be explained by occlusion, yielding a small $e^\perp$. For trajectory B, however, the tree and road in the lower-right region of $x_g$ is inconsistent with the plausible future appearances around $\mu^{(B)}$, where this region is expected to contain buildings, resulting in a large $e^\perp$. Our score thus correctly favors trajectory A.

\begin{figure*}[t]
\centering
\includegraphics[width=\textwidth]{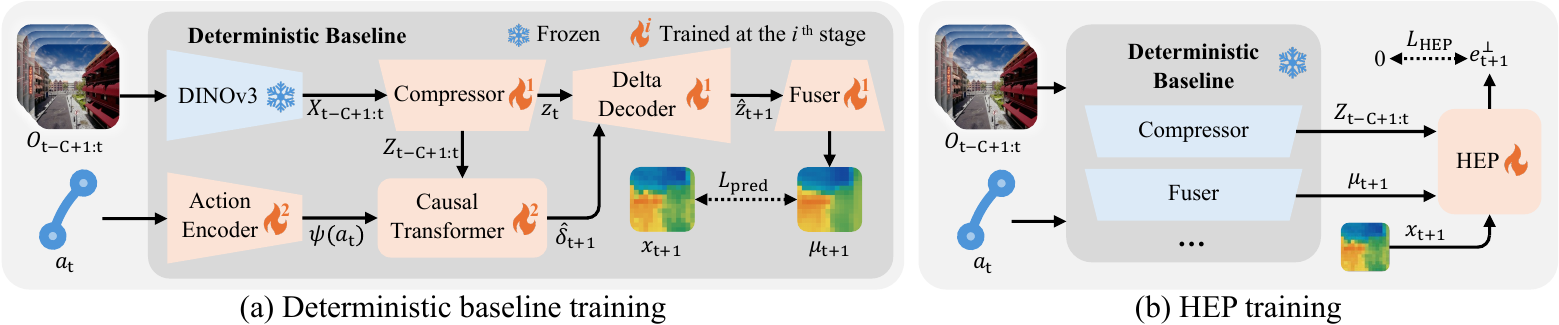}
\caption{The overall training procedure of UA-NWM.}
\label{fig_train}
\end{figure*}

\subsection{Hierarchical Error Projection (HEP)}

To decompose $e$ into $e^\parallel$ and $e^\perp$, we introduce the Hierarchical Error Projection (HEP) module, illustrated in Figure~\ref{fig_hyperplane}(c). HEP takes the deterministic prediction $\mu$ and context observation latents as input and models plausible deviations using multi-scale low-rank subspaces. To capture spatially coherent uncertainty, it decomposes the discrepancy field through a coarse-to-fine pyramid with grid scales
\begin{equation}
    \mathcal{G}=\{1,2,7,14\}.
\end{equation}
At each scale $g$, spatial average pooling partitions the $14\times14$ patch grid of $\mu$ into $g\times g$ cells and yields a descriptor $m_{g,c}$ for each cell $c$. Then a context descriptor $q_{g,c}$ is obtained through cross-attention with context latents, enabling each cell to retrieve relevant contextual evidence. Finally, a scale-specific MLP $f_g$ generates a rank-$R$ basis for each cell:
\begin{equation}
\begin{aligned}
    U_{g,c}
    &=f_g([m_{g,c},q_{g,c}])\\
    &=\{u_{g,c,1},\ldots,u_{g,c,R}\},
    \quad u_{g,c,r}\in \mathbf{R}^{D},
\end{aligned}
\end{equation}
where $[\cdot,\cdot]$ denotes concatenation. 

HEP projects the discrepancy hierarchically from coarse to fine. It maintains a residual field $r$ containing the discrepancy unexplained by coarser scales. Let $r_{g,i}\in\mathbb{R}^{D}$ denote the residual at patch $i$ upon entering scale $g$, initialized as $r_{1,i}=e_i$. At each scale, the residuals within each cell $c$ are averaged:
\begin{equation}
    \bar{r}_{g,c}
    =
    \frac{1}{|\Omega_{g,c}|}
    \sum_{i\in\Omega_{g,c}} r_{g,i} ,
\end{equation}
where $\Omega_{g,c}$ denotes the set of patches in cell $c$. We then project $\bar{r}_{g,c}$ onto the subspace spanned by $U_{g,c}$ by solving a ridge-regularized least-squares problem:
\begin{equation}
    \alpha_{g,c}
    =
    \mathop{\mathrm{argmin}}_{\alpha\in\mathbf{R}^{R}}
    \left\|
    \bar{r}_{g,c}
    -
    \sum_{r=1}^{R}\alpha_r u_{g,c,r}
    \right\|_2^2
    +
    \lambda\|\alpha\|_2^2 ,
\end{equation}
where $\lambda$ is a small regularization coefficient. The resulting uncertainty-explainable component is
\begin{equation}
    p_{g,c}=\sum_{r=1}^{R}\alpha_{g,c,r}u_{g,c,r}.
\end{equation}
This component is subtracted from every patch in the cell to obtain the residual for the next finer scale:
\begin{equation}
    r_{g^+,i} = r_{g,i} - p_{g,c},
    \quad i\in\Omega_{g,c},
\end{equation}
where $g^+$ denotes the next finer scale in $\mathcal{G}$. After processing the finest scale, the remaining residual defines $e^\perp$, while the explained component is given by $e^\parallel=e-e^\perp$. Overall, HEP constructs a coarse-to-fine, region-specific uncertainty subspace that approximates the conditional future-state distribution around the deterministic prediction. This enables uncertainty-aware trajectory scoring in one forward pass without sampling multiple future observations.

\subsection{Deterministic Baseline and Training}

HEP is built upon a deterministic latent world model. A frozen DINOv3~\cite{dinov3} encoder extracts dense features $x_t=\phi(o_t)$, which are compressed into $K$ latent tokens $z_t=C_\eta(x_t)$. Following DeltaWorld~\cite{deltatok}, the transition from $z_t$ to $z_{t+1}$ is represented by $M$ delta tokens that encode their differences. An action-conditioned causal Transformer predicts these delta tokens rather than the next state directly:
\begin{align}
    \hat{\delta}_{t+1}
    &=P_\omega(Z_{t-C+1:t},\psi(a_t)),\\
    \hat{z}_{t+1}
    &=D_\Delta(z_t,\hat{\delta}_{t+1}),\\
    \mu_{t+1}
    &=G_\zeta(\hat{z}_{t+1}),
\end{align}
where $D_\Delta$ is the delta decoder and $G_\zeta$ fuses the predicted compact tokens into the dense DINO feature map $\mu_{t+1}$.

The deterministic baseline is trained in two stages. First, the compressor, fuser, and delta decoder are optimized to learn compact representations and latent transitions. These modules are then frozen, while the action encoder and causal Transformer are trained to predict future latent states. The prediction $\mu_{t+1}$ is directly supervised by the DINO feature of the future observation $x_{t+1}=\phi(o_{t+1})$. Further training details are provided in Section~\ref{sec:training_inference_procedure}.

With the deterministic baseline frozen, HEP is then trained to minimize the fraction of unexplained discrepancy $e^\perp$:
\begin{equation}
    \mathcal{L}_{\mathrm{HEP}}
    =
    \frac{
        \sum_{i=1}^{N}\|e_{t+1,i}^{\perp}\|_2^2
    }{
        \sum_{i=1}^{N}\|e_{t+1,i}\|_2^2+\epsilon
    }.
\end{equation}
The normalized objective prevents large-error samples from dominating training, whereas inference uses the unnormalized orthogonal residual energy for trajectory scoring. The prediction and HEP losses are averaged over an eight-step autoregressive rollout to mitigate error accumulation.

\section{Experiments}

\subsection{Experimental Settings}
\subsubsection{Benchmark} Existing UAV navigation benchmarks mainly target vision-and-language navigation (VLN)~\cite{openuav, uavon} or indoor environments~\cite{sign, flyingimage}, leaving no public benchmark for large-scale outdoor 3D UAV image-goal navigation available. Therefore, we construct a new benchmark named \textbf{AirGoal-10k}, using the AirSim simulator~\cite{airsim} and building upon existing aerial VLN benchmarks~\cite{aerialvln,openfly}. 

In our preliminary analysis, directly cropping short segments from
the original VLN trajectories led to a highly imbalanced turning distribution:
most segments had a heading-change angle below $15^\circ$. Such data would make
the learned world model accurate for nearly straight motion but unreliable under
large turns, which are important for closed-loop navigation. We therefore use
the original trajectory annotations only as feasible start-state pools, and
resample the future motion ourselves to cover a broad range of turn magnitudes
and directions. 

\begin{figure}[t]
\centering
\includegraphics[width=\linewidth]{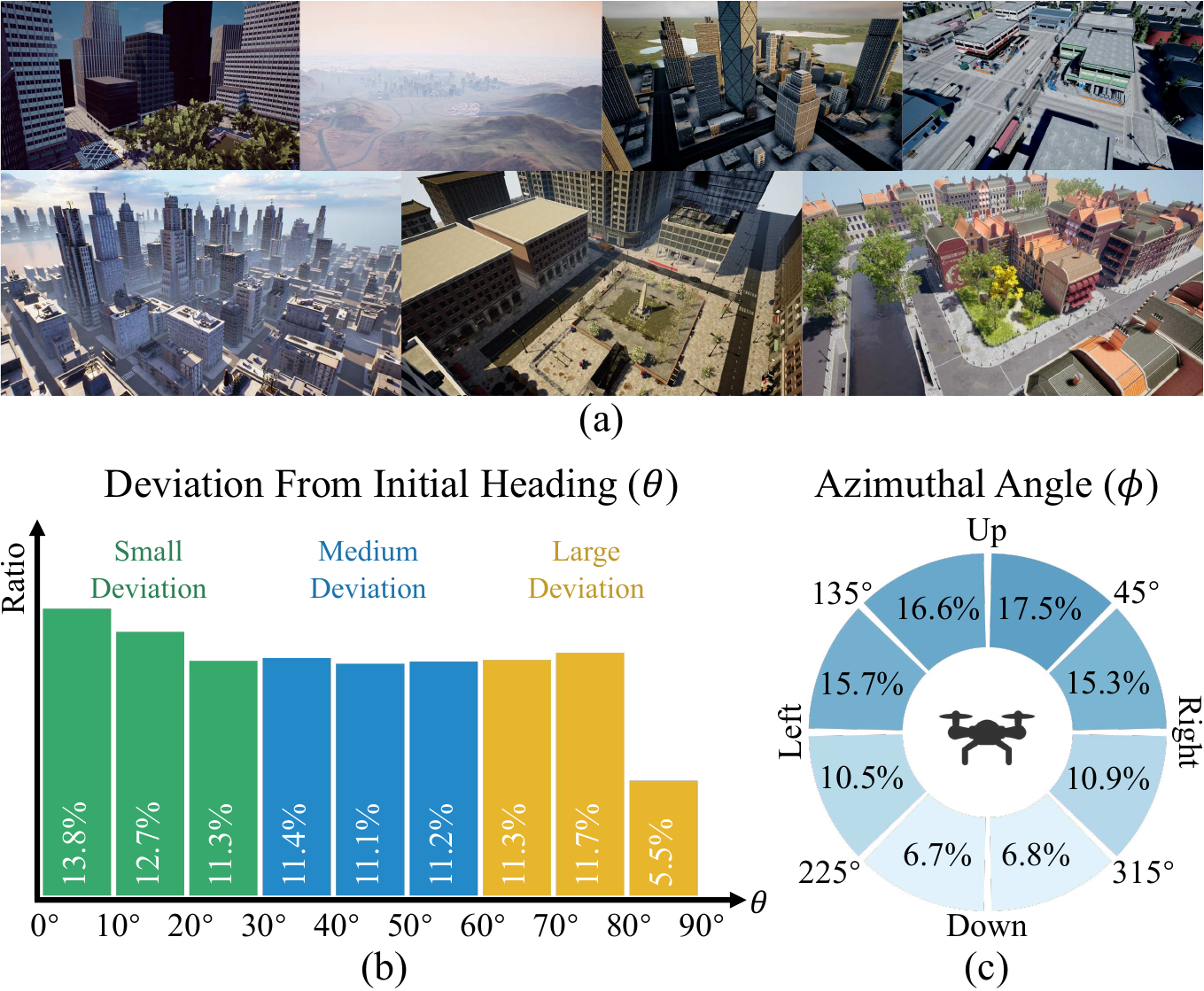}
\caption{Overview of the AirGoal-10k benchmark. (a) Diverse navigation environments. (b) Distribution of deviations from the initial heading. (c) Distribution of azimuthal angles.}
\label{fig_dataset}
\end{figure}

After trajectory sampling, rendering, and quality filtering, AirGoal-10k contains 9000 trajectories for training, 1000 trajectories for validation, and 1000 trajectories for testing. Each trajectory comprises 3D waypoints paired with egocentric RGB observations. As shown in Figure~\ref{fig_dataset}, these trajectories span diverse urban navigation environments. The distribution of the initial-heading deviation $\theta$ is approximately uniform over $[0,90^\circ]$, while the azimuth angle $\varphi$ covers the full range of $[0,360^\circ]$. Together, these distributions preserve diverse left--right and up--down motion components for trajectory-conditioned future prediction. Further details are provided in Section~\ref{sec:dataset_details}.

\begin{table*}[t]
    \centering
    \setlength{\tabcolsep}{1mm}

    \begin{tabularx}{\textwidth}{l*{7}{>{\centering\arraybackslash}X}c}
        \toprule
        \multirow{2}{*}{\raisebox{-0.6ex}{Methods}}
        & \multirow{2}{*}{\raisebox{-0.6ex}{\#Params}}
        & \multicolumn{2}{c}{8 Candidates}
        & \multicolumn{2}{c}{16 Candidates}
        & \multicolumn{2}{c}{32 Candidates}
        & \multirow{2}{*}{\raisebox{-0.6ex}{Time/Frame $\downarrow$}} \\

        \cmidrule(lr){3-4}
        \cmidrule(lr){5-6}
        \cmidrule(lr){7-8}

        & & ATE $\downarrow$ & RPE $\downarrow$
        & ATE $\downarrow$ & RPE $\downarrow$
        & ATE $\downarrow$ & RPE $\downarrow$
        & \\
        \midrule
        
        \textit{Random Selection} 
        & --
        & 1.76 & 0.48
        & 1.79 & 0.49
        & 1.80 & 0.49
        & -- \\
        
        \midrule
        
        NWM~\cite{nwm}
        & 195M
        & 1.47 & 0.40
        & 1.40 & 0.38
        & 1.37 & 0.37
        & 438.67 ms \\

        One-Step WM~\cite{onestepwm}
        & 176M
        & 1.52 & 0.41
        & 1.48 & 0.40
        & 1.44 & 0.39
        & 10.44 ms \\
        
        MWM~\cite{mwm}
        & 202M
        & 1.58 & 0.43
        & 1.60 & 0.43
        & 1.59 & 0.43
        & 122.78 ms \\
        
        RAE-NWM~\cite{raenwm}
        & 223M
        & 1.54 & 0.42
        & 1.54 & 0.42
        & 1.50 & 0.41
        & 386.00 ms \\

        % DreamNav~\cite{pairuav}
        % & \underline{1.32} & \underline{0.37}
        % & \underline{1.28} & \underline{0.36}
        % & \underline{1.25} & \underline{0.35}
        % & 6.83 ms & Arxiv'2026 \\
        
        \midrule

        \rowcolor{gray!20}
        \textbf{Deterministic Baseline (Ours)}
        & 115M
        & \underline{1.39} & \underline{0.38}
        & \underline{1.33} & \underline{0.36}
        & \underline{1.27} & \underline{0.35}
        & 8.06 ms \\
        
        \rowcolor{gray!20}
        \textbf{UA-NWM (Ours)}
        & 122M
        & \textbf{1.26} & \textbf{0.35}
        & \textbf{1.17} & \textbf{0.33}
        & \textbf{1.09} & \textbf{0.30}
        & 8.47 ms \\

        \midrule

        \textit{Oracle Selection} 
        & --
        & 0.90 & 0.26
        & 0.76 & 0.22
        & 0.63 & 0.19
        & -- \\
        
        \bottomrule
    \end{tabularx}

    \caption{Comparison with world-model-based methods on trajectory ranking task of AirGoal-10k.}
    \label{tab:nomad_experiments}
\end{table*}

\subsubsection{Baselines} We compare our method with representative policy-based and world-model-based baselines. The policy-based methods include GNM, ViNT, NoMaD, FlowNav, and NaviBridger~\cite{gnm,vint,nomad,flownav,navibridger}, which directly predict actions or trajectories from visual observations. The world-model-based methods include NWM, One-Step WM, MWM, and RAE-NWM~\cite{nwm,onestepwm,mwm,raenwm}, which predict future observations or latent states to rank candidate trajectories.

\subsubsection{Evaluation Metrics} For offline evaluation, we use Absolute Trajectory Error (ATE) and Relative Pose Error (RPE), following~\cite{nwm}. For online closed-loop evaluation, we report Success Rate (SR), Success weighted by Path Length (SPL), and the average planning time per step, following~\cite{relnwm}. An episode is considered successful if the agent stops within 20 meters of the target position  within the maximum step budget.
\subsubsection{Implementation Details} We use a frozen DINOv3 ViT-B/16 ~\cite{dinov3} to extract dense visual features. Unless otherwise specified, the context length is $C=4$, the autoregressive prediction horizon is $H=8$, and each visual state is compressed into $K=32$ latent tokens. The model is trained with AdamW on four NVIDIA RTX 4090 GPUs. Further details are provided in Section~\ref{sec:implementation_details}.

\begin{table}[t]
    \centering

    \begin{tabularx}{\linewidth}{
        >{\raggedright\arraybackslash}X
        c
        c
        c
    }
        \toprule
        Methods
        & ATE $\downarrow$
        & RPE $\downarrow$
        & VENUE \\
        \midrule

        \multicolumn{4}{c}{\textit{Policy-based Methods}} \\
        \midrule

        GNM~\cite{gnm}
        & 1.29
        & \underline{0.35}
        % & 1.12 ms \\
        & ICRA'2023 \\

        ViNT~\cite{vint}
        & 1.37
        & 0.38
        % & 2.33 ms \\
        & CoRL'2023 \\

        NoMaD~\cite{nomad}
        & 1.79
        & 0.49
        % & 9.94 ms \\
        & ICRA'2024 \\

        FlowNav~\cite{flownav}
        & 1.81
        & 0.50
        % & 9.74 ms \\
        & IROS'2025 \\

        % NaviDiffusor
        % & 1.77
        % & 0.48
        % % & 9.46 ms \\
        % & ICRA'2025 \\

        $\text{NaviBridger}_\text{CVAE}$~\cite{navibridger}
        & 1.58 
        & 0.44 
        % & 34.63 ms \\
        & CVPR'2025 \\

        \midrule
        \multicolumn{4}{c}{\textit{World-model-based Methods}} \\
        \midrule

        NWM~\cite{nwm}
        & 1.40
        & 0.39
        % & 39948.71 ms \\
        % & 39.95 s \\
        & CVPR'2025 \\

        % UniWM
        % & ??
        % & ??
        % & Arxiv'2025 \\

        One-Step WM~\cite{onestepwm}
        & 1.53
        & 0.44
        & Arxiv'2026 \\

        MWM~\cite{mwm}
        & 1.54
        & 0.43
        % & 39933.66 ms \\
        % & 39.93 s \\
        & Arxiv'2026 \\

        RAE-NWM~\cite{raenwm}
        & 1.46
        & 0.43
        % & 6448.26 ms \\
        % & 6.45 s \\
        & ECCV'2026 \\

        % DreamNav
        % & 1.34
        % & 0.36
        % & 6.84 ms \\

        \midrule

        \rowcolor{gray!20}
        \textbf{Det. Baseline (Ours)}
        & \underline{1.26}
        & \underline{0.35}
        % & 26.67 ms \\
        & -- \\

        \rowcolor{gray!20}
        \textbf{UA-NWM (Ours)}
        & \textbf{1.22}
        & \textbf{0.33}
        % & 28.84 ms \\
        & -- \\

        \bottomrule
    \end{tabularx}

    \caption{Comparison with policy-based and world-model-based methods
    on standalone planning task of AirGoal-10k.}
    \label{tab:trajectory_comparison}
\end{table}

\begin{table}[t]
    \centering

    \begin{tabularx}{\linewidth}{
        >{\raggedright\arraybackslash}X
        c
        c
        c
    }
        \toprule
        Methods
        & SR $\uparrow$
        & SPL $\uparrow$
        & Time/Step $\downarrow$ \\
        \midrule

        \multicolumn{4}{c}{\textit{Policy-based Methods}} \\
        \midrule

        GNM~\cite{gnm}
        & 48.0\%
        & 40.9\%
        & 37.46 ms \\

        ViNT~\cite{vint}
        & 53.0\%
        & 46.0\%
        & 48.47 ms \\

        NoMaD~\cite{nomad}
        & 56.0\%
        & 47.2\%
        & 117.49 ms \\

        FlowNav~\cite{flownav}
        & 45.0\%
        & 38.6\%
        & 117.55 ms \\

        % NaviDiffusor
        % & 47.0\%
        % & 47.0\%
        % & 118.80 ms \\

        $\text{NaviBridger}_\text{CVAE}$~\cite{navibridger} & 48.0\% & 41.8\% & 166.08 ms\\

        \midrule
        \multicolumn{4}{c}{\textit{World-model-based Methods}} \\
        \midrule

        NWM~\cite{nwm}
        & 63.0\%
        & 52.1\%
        & 191.71 s \\

        % UniWM
        % & ??
        % & ??
        % & ?? s \\
        
        One-Step WM~\cite{onestepwm}
        & 55.0\%
        & 43.5\%
        & 3.39 s \\

        MWM~\cite{mwm}
        & 69.0\%
        & 56.0\%
        & 66.57 s \\
        % & \textcolor{red}{70.0}\%
        % & \textcolor{red}{69.4}\%
        % & \textcolor{red}{207.59} s \\

        RAE-NWM~\cite{raenwm}
        & \underline{70.0\%}
        & 57.8\%
        & 202.54 s \\
        % & \textcolor{red}{66.0}\%
        % & \textcolor{red}{65.9}\%
        % & \textcolor{red}{38.82} s \\

        % DreamNav
        % & 45.0\%
        % & 44.8\%
        % & 0.03 s \\

        \midrule

        \rowcolor{gray!20}
        \textbf{Det. Baseline (Ours)}
        % & \textcolor{red}{??}\%
        % & \textcolor{red}{??}\%
        & \underline{70.0\%}
        & \underline{60.1\%}
        & 2.38 s \\

        \rowcolor{gray!20}
        \textbf{UA-NWM (Ours)}
        % & \textbf{\textcolor{red}{??}\%}
        % & \textbf{\textcolor{red}{??}\%}
        & \textbf{76.0}\%
        & \textbf{64.5}\%
        & 2.70 s \\

        \bottomrule
    \end{tabularx}

    \caption{Comparison with policy-based and world-model-based methods on online closed-loop navigation task.}
    \label{tab:onlinesimulation}
\end{table}

\subsection{Offline Experiments}
We evaluate two offline settings: trajectory ranking with externally generated candidates and standalone planning without an external policy. For trajectory ranking, NoMaD~\cite{nomad} generates 8, 16, or 32 candidate action sequences, which are then scored and selected by each world-model-based method using identical candidate sets for fairness. As shown in Table~\ref{tab:nomad_experiments}, UA-NWM achieves the lowest ATE and RPE across all candidate-set sizes while remaining substantially faster than previous methods, especially stochastic models requiring iterative generation. For standalone planning, policy-based methods directly predict one trajectory, whereas world models optimize action sequences with CEM~\cite{cem} following NWM~\cite{nwm}. At each of three iterations, CEM samples and scores 32 candidates and updates the sampling distribution using the top 16 elites, after which the final distribution mean is used as the planned trajectory. As shown in Table~\ref{tab:trajectory_comparison}, UA-NWM still achieves the best planning performance.

\subsection{Online Simulation Experiments}
We evaluate closed-loop navigation on 100 episodes in AirSim~\cite{airsim}. In each episode, the agent navigates to the location specified by a goal image with unknown goal distance. At each step, policy-based methods directly predict a trajectory, whereas world models plan with CEM~\cite{cem}. As shown in Table~\ref{tab:onlinesimulation}, UA-NWM achieves the best online performance. Compared with offline planning, this setting introduces a distribution shift from fixed-length trajectories to variable-length, long-horizon navigation. Policy-based methods degrade substantially, while autoregressive world models generalize more robustly.

\subsection{Qualitative Analysis}

\begin{figure*}[t]
\centering
\includegraphics[width=\textwidth]{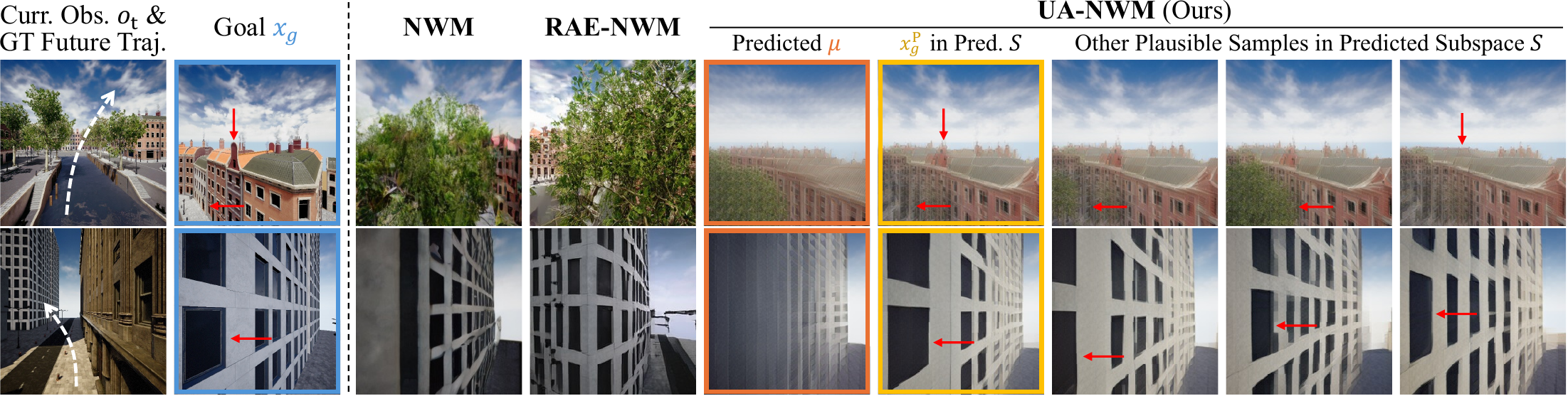}
\caption{Qualitative comparison of future predictions. The DINO latent features predicted by RAE-NWM and UA-NWM are decoded using RAE~\cite{rae} and RAEv2~\cite{raev2}, respectively, solely for visualization. RAE-NWM and UA-NWM perform trajectory scoring directly in the latent space, whereas NWM scores LPIPS~\cite{lpips} of predicted RGB images. Details for latent visualization are provided in Section~\ref{sec:visualization_details}.}
\label{fig_visual_compare}
\end{figure*}

Figure~\ref{fig_visual_compare} compares predictions under ground-truth trajectories. NWM~\cite{nwm} and RAE-NWM~\cite{raenwm} exhibit substantial error accumulation during multi-step autoregressive rollout.

For UA-NWM, the deterministic backbone predicts a blurry mean future latent $\mu$, which may deviate considerably from $x_g$ when multiple futures are plausible. HEP decomposes their residual
$e$. The projection $x_g^P=\mu+e^\parallel=x_g-e^\perp$ is the point in the subspace $\mathcal{S}$ closest to $x_g$. Its remaining discrepancy from $x_g$ is exactly $e^\perp$, which is substantially smaller than the original $e$, and will be used for trajectory scoring.

Sampling additional points from $\mathcal{S}$ yields various plausible futures, demonstrating that it captures meaningful variations beyond a single deterministic prediction, which further reveal two major sources of \textit{future-state uncertainty} in navigation:

\begin{itemize}
\item \textbf{Occlusion-induced ambiguity.}
In the first example, foreground trees limit the initial observation, leaving both tree coverage and roof structure in future observation uncertain. The goal image represents one plausible future with no trees in view and a curvilinear gable on the roof. While $\mu$ averages over these possibilities, $x_g^P$ recovers this goal-consistent state. Other samples show varying tree coverage or a roof without the gable, all plausible given the limited initial evidence.

\item \textbf{Long-horizon drift.}
Small pose errors accumulate over long rollouts, making fine-grained details uncertain even without occlusion. In the second example, the overall building layout remains predictable, whereas window positions vary substantially with minor trajectory drift. Modeling this uncertainty too broadly may absorb true trajectory errors and weaken scoring, as seen in NWM~\cite{nwm} and RAE-NWM~\cite{raenwm}. UA-NWM instead preserves global geometry while modeling plausible local variations, balancing uncertainty tolerance and trajectory discrimination.
\end{itemize}

Additional qualitative results are provided in Section~\ref{sec:additional_visualization}.

\begin{table}[t]
    \centering
    \setlength{\tabcolsep}{3pt}

    \begin{tabularx}{\linewidth}{
        *{4}{
            >{\centering\arraybackslash
              \hsize=0.67\hsize
              \linewidth=\hsize}X
        }
        *{4}{
            >{\centering\arraybackslash
              \hsize=1.33\hsize
              \linewidth=\hsize}X
        }
    }
        \toprule
        \multicolumn{4}{c}{Scales $g$}
        & \multicolumn{2}{c}{16 Candidates}
        & \multicolumn{2}{c}{32 Candidates} \\

        \cmidrule(lr){1-4}
        \cmidrule(lr){5-6}
        \cmidrule(lr){7-8}

        1 & 2 & 7 & 14
        & \mbox{ATE $\downarrow$}
        & \mbox{RPE $\downarrow$}
        & \mbox{ATE $\downarrow$}
        & \mbox{RPE $\downarrow$} \\
        \midrule

        $\times$ & $\times$ & $\times$ & $\checkmark$
        & 1.22 & 0.34
        & 1.15 & 0.32 \\

        $\times$ & $\checkmark$ & $\checkmark$ & $\checkmark$
        & 1.20 & \textbf{0.33}
        & 1.12 & 0.31 \\

        $\checkmark$ & $\times$ & $\checkmark$ & $\checkmark$
        & 1.21 & 0.34
        & 1.13 & 0.31 \\

        $\checkmark$ & $\checkmark$ & $\times$ & $\checkmark$
        & 1.21 & 0.34
        & 1.14 & 0.32 \\

        $\checkmark$ & $\checkmark$ & $\checkmark$ & $\times$
        & 1.24 & 0.34
        & 1.17 & 0.32 \\

        % \midrule
        \rowcolor{gray!20}
        $\checkmark$ & $\checkmark$ & $\checkmark$ & $\checkmark$
        & \textbf{1.17} & \textbf{0.33}
        & \textbf{1.09} & \textbf{0.30} \\

        \bottomrule
    \end{tabularx}

    \caption{Ablation on HEP's components of different scales.}
    \label{tab:scale_ablation}
\end{table}

\begin{table}[t]
    \centering
    \setlength{\tabcolsep}{3pt}

    \begin{tabularx}{\linewidth}{
        c
        *{4}{
            >{\centering\arraybackslash}X
        }
    }
        \toprule
        \multirow{2}{*}{\raisebox{-0.6ex}{Rank $R$}}
        & \multicolumn{2}{c}{16 Candidates}
        & \multicolumn{2}{c}{32 Candidates} \\

        \cmidrule(lr){2-3}
        \cmidrule(lr){4-5}

        & \mbox{ATE $\downarrow$}
        & \mbox{RPE $\downarrow$}
        & \mbox{ATE $\downarrow$}
        & \mbox{RPE $\downarrow$} \\
        \midrule

        1
        & 1.23 & 0.34
        & 1.16 & 0.32 \\
        
        \rowcolor{gray!20}
        2
        & \textbf{1.17} & \textbf{0.33}
        & \textbf{1.09} & \textbf{0.30} \\

        3
        & 1.20 & \textbf{0.33}
        & 1.13 & 0.31 \\

        4
        & 1.21 & 0.34
        & 1.14 & 0.32 \\

        5
        & 1.21 & \textbf{0.33}
        & 1.14 & 0.32 \\

        \bottomrule
    \end{tabularx}

    \caption{Ablation on different values of the basis rank $R$.}
    \label{tab:rank_ablation}
\end{table}

\begin{figure}[t]
\centering
\includegraphics[width=\linewidth]{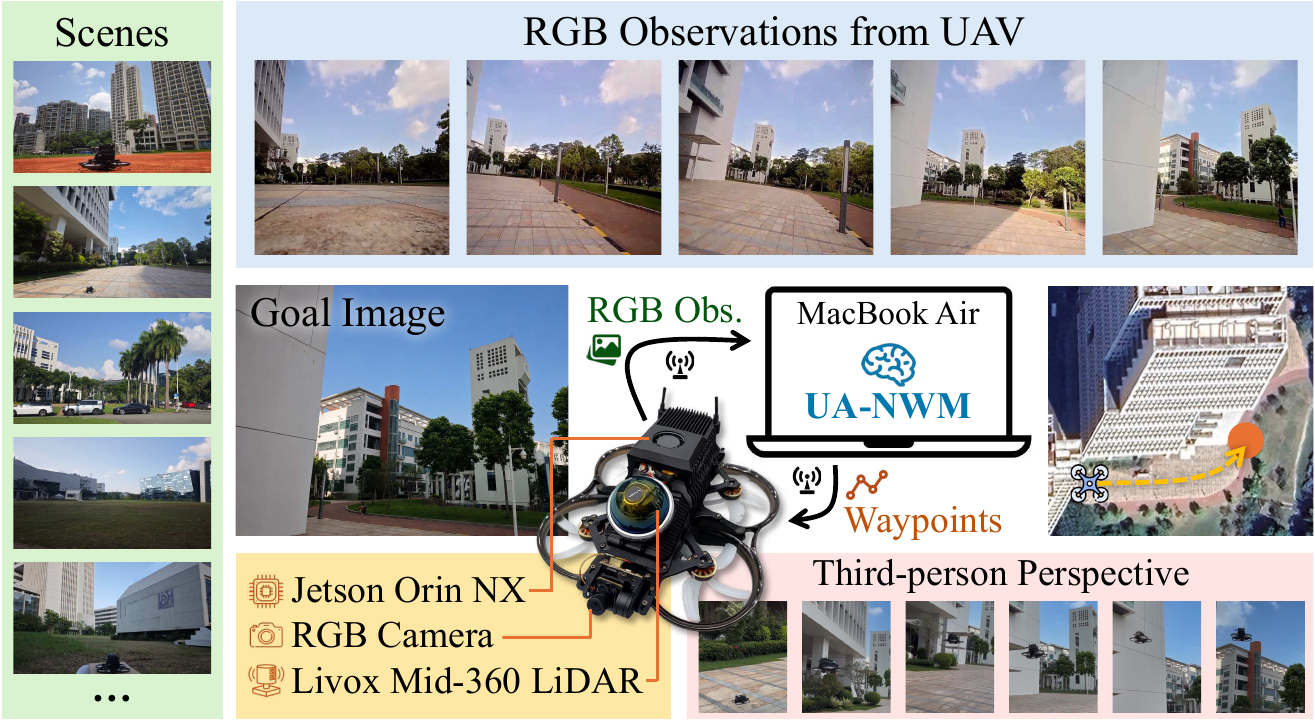}
\caption{Illustration of real-world UAV experiments.}
\label{fig_real}
\end{figure}

\subsection{Ablation Studies}

\subsubsection{Hierarchical Structure} We ablate each scale of HEP to assess its contribution. As shown in Table~\ref{tab:scale_ablation}, removing any scale degrades performance. Scale 14 contributes the most, suggesting that explainable errors $e^\parallel$ predominantly arise from fine-grained local variations, whereas Scale 1 contributes the least because such variations rarely span the entire image.

\subsubsection{Rank of the Basis} We vary the basis rank $R$ across all scales. As shown in Table~\ref{tab:rank_ablation}, $R=2$ performs best, suggesting that a low-dimensional subspace is sufficient to capture the dominant directions of $e^\parallel$. A smaller rank underrepresents future uncertainty, whereas a larger rank may absorb unexplainable errors and weaken trajectory discrimination.

\subsection{Real-world Experiments} To evaluate real-world applicability, we deploy UA-NWM locally on a battery-powered MacBook Air and test it on a custom-built quadrotor, as shown in Figure~\ref{fig_real}. The MacBook communicates with the UAV via a mobile hotspot, performs onboard planning, and transmits the predicted 3D waypoints to the low-level flight controller for execution. At each step, UA-NWM plans an 8-waypoint trajectory using CEM~\cite{cem}, executes the first waypoint, and replans from the latest observation. On-device planning takes approximately 9 seconds per trajectory, including communication latency. Experiments on five navigation tasks across multiple scenes demonstrate successful closed-loop deployment and zero-shot sim-to-real generalization without fine-tuning on real-world data. Details are provided in Section~\ref{sec:real_world_details}.

% \subsubsection{Ablation Studies}

\section{Conclusion}

We presented UA-NWM, an uncertainty-aware world model for aerial image-goal navigation. UA-NWM formulates trajectory scoring as conditional OOD detection, models plausible future variations through an uncertainty subspace, and separates plausible uncertainty-induced deviations from unexplained residual errors. This enables efficient distribution-aware trajectory scoring without stochastic future sampling. Experiments show that UA-NWM improves navigation performance across diverse tasks, while preserving low inference latency, and real-world UAV deployment further supports its practical applicability.

%% file: sec/X_suppl.tex
\clearpage
\maketitlesupplementary

\appendix
\tableofcontents

\section{Additional Related Work}

\subsection{Uncertainty-Aware Visual Navigation}

Uncertainty in visual navigation has been studied primarily in current perception and semantic representation. Existing methods calibrate and temporally aggregate segmentation predictions for object navigation~\cite{perceptionmatters}, use vision--language uncertainty to guide active object search~\cite{uncertaintyactive}, or integrate geometric, semantic, and appearance uncertainty into 3D Gaussian maps for vision-language navigation~\cite{uagm}. Others estimate model familiarity or action confidence to detect out-of-distribution observations, invoke additional reasoning, or request assistance~\cite{competencynav,adanav,interactivenav}.

A complementary line of work considers uncertainty in localization, mapping, and motion planning. Perception-aware planners select trajectories that reduce visual localization uncertainty~\cite{perceptionaware}, while UrbanFly~\cite{urbanfly} models uncertainty in monocular visual--inertial maps for collision-aware UAV planning. Other approaches propagate state-estimation and model uncertainty into collision prediction and action selection~\cite{uncertaintyattention}, or construct calibrated uncertainty sets from noisy visual predictions for robust task-level optimization~\cite{neuro}.

These methods primarily address uncertainty in current perception, localization, semantic mapping, or planning constraints. In contrast, we focus on \textit{future-state uncertainty}: given the same visual context and candidate trajectory, multiple future observations may be plausible.

\subsection{OOD Detection} Out-of-distribution (OOD) detection aims to identify test samples that deviate from the distribution represented by a model. Conventional approaches derive OOD scores from classifier confidence, energy functions, or distances in a learned feature space \cite{hendrycks2016baseline,lee2018simple,liu2020energy}. Generative approaches instead use likelihood, reconstruction error, or predictive uncertainty to determine whether an observation is supported by the learned data distribution \cite{nalisnick2018deep,ren2019likelihood}. However, these methods are primarily designed to detect distribution shifts in static inputs and do not account for how future observations depend on an agent's context and actions.

Recent studies have incorporated uncertainty and OOD detection into world models to identify unreliable predictions or unseen hazards \cite{seo2025uncertainty,mei2025c3}. Their primary objective is to determine whether a world model is operating outside its training distribution, rather than to evaluate whether a visual goal is compatible with the future induced by a particular candidate trajectory.

\section{Implementation Details of UA-NWM}
\label{sec:implementation_details}
This section provides further details on the architecture, training and inference procedures, and hyperparameter settings of UA-NWM.

\subsection{Model Structure}
\label{sec:model_structure}
UA-NWM consists of a lightweight deterministic latent world model and the proposed Hierarchical Error Projection (HEP) module. Their overall architectures are illustrated in Figure~\ref{fig_hyperplane}(c) and Figure~\ref{fig_train}(a), respectively. We provide the detailed structure of each component below.

\paragraph{DINO encoder}
Each RGB observation is resized to $224\times224$ and encoded by a frozen DINOv3 ViT-B/16 encoder~\cite{dinov3}. We discard the class and register tokens and retain the
$14 \times 14$ patch grid, yielding $N=196$ patch tokens with
feature dimension $D=768$:
\begin{equation}
    x_t=\phi(o_t)\in\mathbb{R}^{N\times D}.
\end{equation}
\paragraph{Token compressor.}
The compressor maps the dense DINO feature representation into $K=32$ compact latent tokens, each with dimension $d=384$:
\begin{equation}
    z_t=C_\eta(x_t)\in\mathbb{R}^{K\times d}.
\end{equation}
It comprises $K$ learnable queries of dimension 384 and a six-head cross-attention layer. The normalized 768-dimensional DINO patch tokens are used as keys and values and are projected to the 384-dimensional attention space inside the cross-attention layer. Let $Q_C\in\mathbb{R}^{K\times 384}$ denote the learned compressor queries. The compression process is formulated as
\begin{align}
    h_t^C &= \operatorname{MHA}_C
    \bigl(\operatorname{LN}(Q_C),
    \operatorname{LN}(x_t),\operatorname{LN}(x_t)\bigr),\\
    z_t &= h_t^C+\operatorname{MLP}_C
    \bigl(\operatorname{LN}(h_t^C)\bigr),
\end{align}
where the three inputs to $\operatorname{MHA}_C$ correspond to the queries, keys, and values, respectively. The feed-forward network $\operatorname{MLP}_C$ consists of
$\operatorname{Linear}(384,384)$, GELU, and
$\operatorname{Linear}(384,384)$. All dropout rates in the token compressor are set to zero.

\paragraph{Delta encoder and decoder.}
Following the delta-aware transition parameterization of DeltaWorld~\cite{deltatok}, we use $M=32$ delta tokens, each with dimension 384, to represent the transition between two adjacent frames. During representation training, a teacher delta encoder extracts the latent transition from two adjacent compressed states. For each token slot, it concatenates the current state, next state, and their discrepancy,
\begin{equation}
    u_{t+1}=\left[z_t;z_{t+1};z_{t+1}-z_t\right]
    \in\mathbb{R}^{K\times 3d},
\end{equation}
and projects $u_{t+1}$ from 1152 to 384 dimensions. The projected tokens are processed by two pre-norm Transformer encoder layers, each containing 6-head self-attention and an MLP with dimensions $384\rightarrow1536\rightarrow384$ and GELU activation. A final LayerNorm and linear projection produce
\begin{equation}
    \delta_{t+1}=E_\Delta(z_t,z_{t+1})
    \in\mathbb{R}^{32\times384}.
\end{equation}
The delta encoder is used only to provide teacher delta tokens during representation training and is not required at inference.

The delta decoder has the same two-layer Transformer structure. It first concatenates $z_t$ and a delta latent token-wise and projects the resulting 768-dimensional features to 384 dimensions. After two pre-norm Transformer layers and a final LayerNorm-linear projection, it predicts a residual update that is added to the current state:
\begin{equation}
    D_\Delta(z_t,\delta)=z_t+
    W_\Delta\operatorname{LN}\bigl(
    \operatorname{Tr}_\Delta([z_t;\delta])\bigr).
\end{equation}
Here, $\delta\in\mathbb{R}^{M\times384}$ denotes the input delta tokens, $[\cdot;\cdot]$ denotes token-wise feature concatenation, $\operatorname{Tr}_\Delta$ denotes the Transformer decoder, and $W_\Delta$ is the final linear projection.

\paragraph{Action-conditioned causal transformer.}
Given the compressed context states and a future action sequence, UA-NWM autoregressively applies a one-step transition at each action step. The normalized four-dimensional action $a_t$ is embedded as
\begin{equation}
    \psi(a_t)=W_2\operatorname{SiLU}(W_1a_t)\in\mathbb{R}^{384},
\end{equation}
where $W_1$ and $W_2$ denote linear projections, and both the hidden and output dimensions are 384. The predictor receives $C=4$ compressed context states. Learned frame and token positional embeddings are added before the $4\times32=128$ context tokens are flattened and processed by six causal Transformer blocks. Each block contains 6-head self-attention and an MLP with dimensions $384\rightarrow1536\rightarrow384$ and GELU activation. The attention mask is causal across frames while allowing all tokens within the same frame to interact.

The action embedding modulates every Transformer block through adaptive LayerNorm (AdaLN). Each block uses an independent SiLU-linear modulation layer that maps the 384-dimensional action embedding to six 384-dimensional vectors: shift, scale, and residual gate parameters for the attention and MLP branches. The modulation output layer is zero-initialized. After the six blocks, only the 32 tokens associated with the most valuable information are retained and passed through a final LayerNorm and linear head to predict the delta latent:
\begin{equation}
    \hat\delta_{t+1}
    =P_\omega\bigl(Z_{t-C+1:t},\psi(a_t)\bigr)
    \in\mathbb{R}^{32\times384}.
\end{equation}
The next absolute latent state is then composed by the delta decoder,
\begin{equation}
    \hat z_{t+1}=D_\Delta(z_t,\hat\delta_{t+1}).
\end{equation}
The predicted state is appended to the context and the oldest state is removed before processing the next action, yielding an autoregressive rollout over $A_{t:t+H-1}$.

\paragraph{Token fuser.}
The token fuser decodes a predicted compressed state into a dense DINO-space feature map. It starts from 196 learned 384-dimensional output queries and applies two cross-attention blocks over the 32 compressed tokens. Each block contains pre-normalized 6-head cross-attention, a residual connection, and a pre-normalized MLP with dimensions $384\rightarrow1536\rightarrow384$ and GELU activation. A final LayerNorm produces the shared per-patch hidden field
$h_{t+1}\in\mathbb{R}^{196\times384}$, and a linear readout maps it to the deterministic mean prediction
\begin{equation}
    \mu_{t+1}=G_\zeta(\hat z_{t+1})
    \in\mathbb{R}^{196\times768}.
\end{equation}
The deterministic baseline scores a candidate trajectory using the flattened cosine distance between its final-step mean prediction and the DINO feature map of the goal image. HEP uses the same output from the fuser , ensuring that the deterministic and uncertainty-aware variants share an identical predictive backbone.

\paragraph{Hierarchical Error Projection module.}
HEP predicts context- and trajectory-conditioned directions along which the true future feature may plausibly deviate from the deterministic prediction. Its basis predictor takes the fuser output and the compressed observation context
$Z_{\mathrm{ctx}}\in\mathbb{R}^{(CK)\times d}$ as input, where $C=4$, $K=32$, and thus $CK=128$. 

The goal feature is not involved in basis prediction; it is introduced only when the goal--prediction discrepancy is projected onto the subspace spanned by the predicted bases. Therefore, the uncertainty subspace is determined solely by the observation context and candidate trajectory through the rolled-out latent state, independently of the goal image.

HEP operates over a spatial pyramid. At scale $g$, the $14\times14$ patch grid is partitioned into $g^2$ cells and average-pooled within each cell, producing a cell descriptor
\begin{equation}
    m_{g,c}=\frac{1}{|\Omega_{g,c}|}
    \sum_{i\in\Omega_{g,c}}h_i
    \in\mathbb{R}^{384}.
\end{equation}
The four scales contain $1$, $4$, $49$, and $196$ cells, corresponding to $196$, $49$, $4$, and $1$ DINO patches per cell, respectively.

Each scale has an independent 4-head context cross-attention layer. The cell descriptors are used as queries, while the full 128 compressed context tokens are used as keys and values:
\begin{equation}
    q_{g,c}=\operatorname{MHA}_g
    \bigl(\operatorname{LN}_g(m_{g,c}),
    \operatorname{LN}(Z_{\mathrm{ctx}}),
    \operatorname{LN}(Z_{\mathrm{ctx}})\bigr)
    \in\mathbb{R}^{384}.
\end{equation}
The cell descriptor and attended context descriptor are concatenated and passed through a scale-specific basis MLP,
\begin{equation}
    U_{g,c}=f_g([m_{g,c},q_{g,c}])
    \in\mathbb{R}^{D\times R},
\end{equation}
where $R=2$, $D=768$, and each $f_g$ has the structure $768\rightarrow384\rightarrow384\rightarrow1536$ with GELU after the first two linear layers. The final 1536-dimensional output represents two 768-dimensional basis directions. The four scales use separate attention and MLP parameters.

Given the normalized goal discrepancy $e$, HEP performs a one-pass coarse-to-fine residual decomposition. Let $r$ denote the residual entering a scale, initialized as $r=e$. For each cell, HEP computes the cell-average residual $\bar r_{g,c}$ and obtains ridge-regularized projection coefficients
\begin{equation}
    \alpha_{g,c}=
    \left(U_{g,c}^{\top}U_{g,c}+\lambda I\right)^{-1}
    U_{g,c}^{\top}\bar r_{g,c},
\end{equation}
where $\lambda=10^{-3}$. The explained component $p_{g,c}=U_{g,c}\alpha_{g,c}$ is subtracted from every patch residual in the cell. The resulting residual is then passed to the next finer scale. Coarse scales remove globally or regionally coherent deviations, whereas the finest $14\times14$ scale performs patch-wise low-rank projection. The basis vectors are not explicitly orthogonalized; the Gram-matrix solve above projects onto their span and remains stable when the predicted directions are correlated.

After the finest scale, the remaining field is the hierarchically unexplained residual $e^\perp$.  This procedure is a sequential ridge projection over the scale pyramid, rather than a single global orthogonal projection onto the union of all scale-wise subspaces.

% \paragraph{Parameter counts.}
% The frozen DINOv3 encoder contains 85.66M parameters. Excluding DINOv3 and HEP, the deterministic model contains 29.52M parameters when the teacher-only delta encoder is included, or 25.38M parameters for the inference-time backbone. HEP adds 6.51M parameters, resulting in 36.03M non-DINO parameters for the complete model.

\begin{figure*}[t]
\centering
\includegraphics[width=\textwidth]{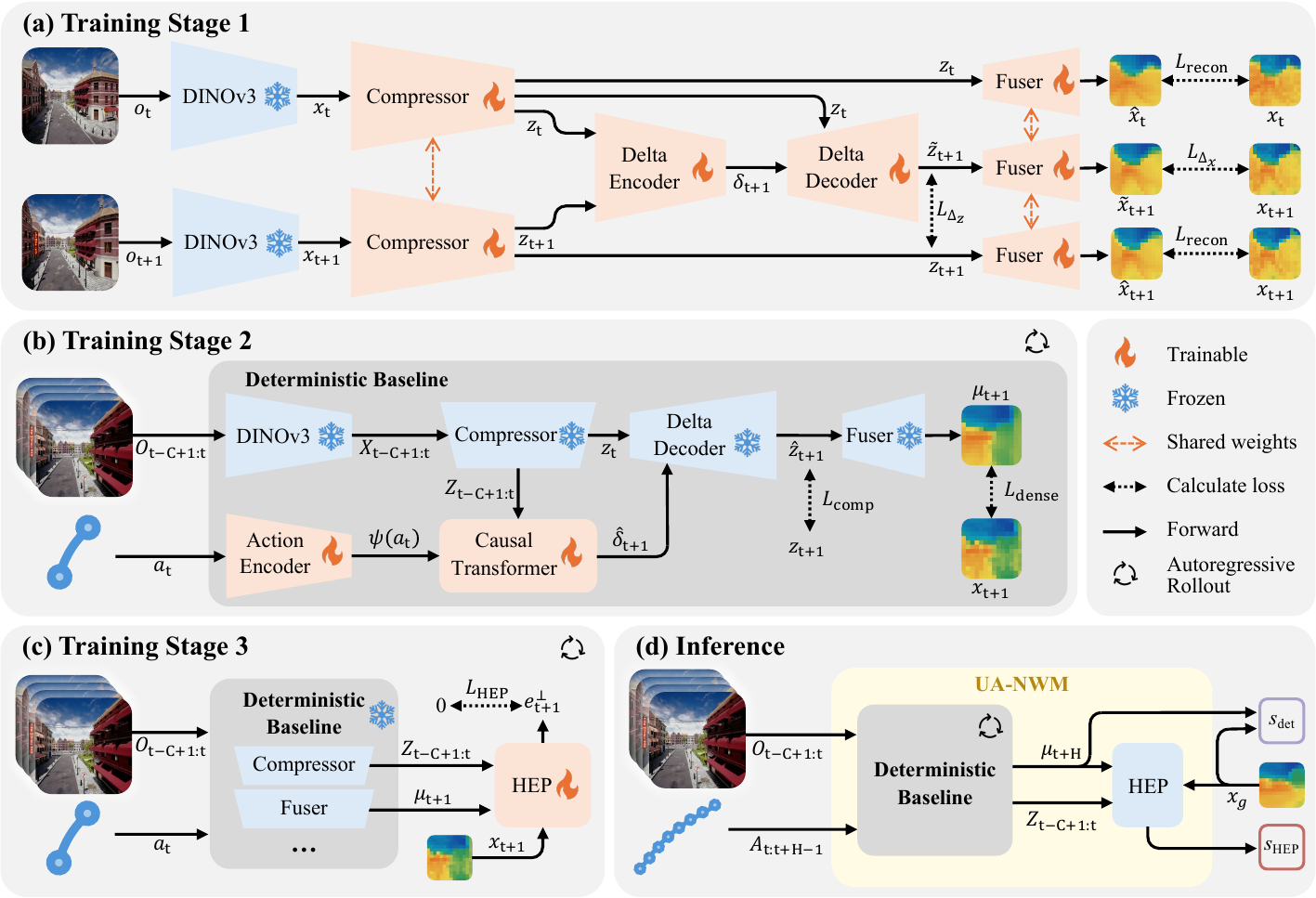}
\caption{Detailed training and inference procedure of UA-NWM.}
\label{fig_backbone}
\end{figure*}

\subsection{Training and Inference Procedure}
\label{sec:training_inference_procedure}

UA-NWM is trained in three stages, as illustrated in Figure~\ref{fig_backbone}. The first two stages learn the deterministic latent world model baseline, and the final stage learns the uncertainty-aware HEP module. 

\paragraph{Stage 1: representation training.}
We first train the representation modules, including the compressor, fuser, delta encoder, and delta decoder. Given an input feature map $x_t$, the compressor produces $z_t=C_\eta(x_t)$, and the fuser reconstructs $\hat{x}_t=G_\zeta(z_t)$. The basic reconstruction objective is
\begin{equation}
    \mathcal{L}_{\mathrm{recon}}
    =
    \left\|\hat{x}_t-x_t\right\|_2^2
    +
    \beta_{\mathrm{rec}} \left(1-\cos(\hat{x}_t,x_t)\right).
\end{equation}
This stage also trains the delta encoder and decoder. Given two adjacent compressed states $z_t$ and $z_{t+1}$, the delta encoder extracts transition latent $\delta_{t+1}$, and the delta decoder reconstructs the next compressed state:
\begin{equation}
    \delta_{t+1}=E_\Delta(z_t,z_{t+1}),
    \quad
    \tilde{z}_{t+1}=D_\Delta(z_t,\delta_{t+1}).
\end{equation}
We decode the reconstructed state as $\tilde{x}_{t+1}=G_\zeta(\tilde{z}_{t+1})$.
This transition is supervised in both compressed latent space and dense DINO space:
\begin{equation}
    \mathcal{L}_{\Delta z}
    =
    \left\|\tilde{z}_{t+1}-z_{t+1}\right\|_2^2,
\end{equation}
\begin{equation}
    \mathcal{L}_{\Delta x}
    =
    \left\|\tilde{x}_{t+1}-x_{t+1}\right\|_2^2
    +
    \rho\left(1-\cos(\tilde{x}_{t+1},x_{t+1})\right).
\end{equation}
The objective of stage 1 is:
\begin{equation}
\mathcal{L}_{\mathrm{rep}}=\gamma_1\cdot\mathcal{L}_{\mathrm{recon}}+\gamma_2\cdot\mathcal{L}_{\Delta z}+\gamma_3\cdot\mathcal{L}_{\Delta x},
\end{equation}
where $\gamma_1$, $\gamma_2$ and $\gamma_3$ are scalar weights applied to each term in implementation.

\paragraph{Stage 2: deterministic prediction training.}
We then train the action encoder and action-conditioned causal Transformer with the compressor, fuser and delta decoder fixed. The prediction is supervised by the DINO feature and compressed latent of the target:
\begin{equation}
    \mathcal{L}_{\mathrm{dense}}
    =
    \left\|\mu_{t+1}-x_{t+1}\right\|_2^2
    +
    \beta_{\mathrm{pred}} \left(1-\cos(\mu_{t+1},x_{t+1})\right),
\end{equation}
\begin{equation}
    \mathcal{L}_{\mathrm{comp}}
    = \lambda_{\mathrm{comp}}\left\|\hat{z}_{t+1}-z_{t+1}\right\|_2^2 ,
\end{equation}
\begin{equation}
    \mathcal{L}_{\mathrm{pred}}
    =
    \mathcal{L}_{\mathrm{dense}}
    +
    \mathcal{L}_{\mathrm{comp}}.
\end{equation}

\paragraph{Stage 3: HEP training.}
Finally, we freeze the deterministic baseline and train only HEP. Given the predicted dense feature $\mu_{t+1}$ and target feature $x_{t+1}$, HEP predicts the uncertainty subspace and computes the orthogonal residual $e_{t+1}^\perp$. The objective minimizes the normalized orthogonal residual fraction:
\begin{equation}
    \mathcal{L}_{\mathrm{HEP}}
    =
    \frac{
    \sum_{i=1}^{N}\left\|e_{t+1,i}^{\perp}\right\|_2^2
    }{
    \sum_{i=1}^{N}\left\|e_{t+1,i}\right\|_2^2+\epsilon
    } .
\end{equation}
This loss is magnitude-invariant and encourages HEP to learn directions along which the true future may plausibly deviate from the deterministic mean.

Although the objectives above are written for a single transition for clarity, UA-NWM uses an autoregressive eight-step rollout for both stage 2 and stage 3 training in order to mitigate error accumulation. The same losses are applied to every predicted step and averaged over time.

During training, we enable short-context augmentation. When a sampled window is close to the beginning of a trajectory with fewer than $C=4$ real context frames available, the missing front context slots are filled by repeating the earliest available frame. The corresponding repeated-frame action intervals become zero-motion actions. This exposes the model to deficient context windows that may occur at the beginning of closed-loop simulation or deployment.

\paragraph{Inference.}
At inference time, UA-NWM consists of the deterministic latent baseline as backbone
and the HEP module as a scoring head. Given a candidate action sequence
$A_{t:t+H-1}$, the backbone first encodes the observation context into
$Z_{t-C+1:t}$ and autoregressively predicts future compressed states. At each rollout step,
the predictor estimates a delta latent from the current latent context and the
corresponding action window; the delta decoder composes the next compressed
state, which is then appended to the context for the following step.
After $H$ steps, the rollout returns predicted states
$\hat z_{t+1:t+H}$. Candidate selection uses only the final state, whose fuser
output is the final-horizon dense DINO prediction $\mu_{t+H}=G_\zeta(\hat z_{t+H})$, because the goal image specifies the desired
terminal observation.

The deterministic baseline and UA-NWM share the same backbone and
differ only in the scoring mechanism. The deterministic baseline directly
compares the mean prediction with the goal DINO feature using flattened
cosine distance:
\begin{equation}
    s_{\mathrm{det}}
    =
    1-\cos\!\left(
    \mathrm{vec}(\mu_{t+H}),
    \mathrm{vec}(x_g)
    \right).
\end{equation}
UA-NWM instead applies HEP as a final-step scoring head after the autoregressive rollout has finished. HEP predicts the uncertainty subspace
from the deterministic prediction and the compressed context tokens;
the goal feature is used only when projecting the final discrepancy, and the HEP
output is not fed back into the rollout dynamics. Let
\begin{equation}
    e_{t+H}=\mathrm{norm}(x_g)-\mathrm{norm}(\mu_{t+H}),
\end{equation}
and let $e_{t+H}^{\perp}$ be the residual left by the HEP decomposition. The
trajectory cost is the unnormalized residual energy:
\begin{equation}
    s_{\mathrm{HEP}}
    =
    \frac{1}{N}
    \sum_{i=1}^{N}
    \left\|e_{t+H,i}^{\perp}\right\|_2^2 .
\end{equation}
This differs from the normalized training loss: at test time, retaining the
absolute magnitude gives sharper discrimination between candidate
trajectories. In offline ranking, the candidate with the lowest score is
selected from the fixed candidate set. In CEM planning, the same score is used as
the optimization cost for sampled action sequences.

\subsection{Hyper-parameter Settings}

Unless otherwise stated, all training stages use AdamW with
$\beta_1=0.9$, $\beta_2=0.95$, bf16 automatic mixed precision, gradient clipping
with norm $1.0$, and a linear warmup followed by cosine decay. The action
representation is a 4-dimensional local-frame delta pose. Position deltas are
normalized by the waypoint spacing of $5.0$ meters, and the action
normalizer is reused for both training and evaluation.

For training stage 1, we use a per-GPU batch size of 12 under 4-GPU DDP,
giving an effective batch size of 48. Stage 1 is trained for 10k steps with
learning rate $10^{-4}$, weight decay $0.05$, 600 warmup steps, and minimum
learning-rate ratio $0.01$. The loss weights are
$\beta_{\mathrm{rec}}=0.05$, $\rho=0.2$, and
$\gamma_1=\gamma_2=\gamma_3=1.0$. We inject Gaussian noise with
standard deviation $0.01$ into compressed latents and $0.005$ into teacher delta
tokens during this stage.

For training stage 2, the rollout horizon is 8. The predictor
is trained for 24k steps with learning rate $10^{-4}$, weight decay $0.05$, 1000
warmup steps, and minimum learning-rate ratio $0.01$. The dense DINO loss uses
$\beta_{\mathrm{pred}}=0.05$, and the compressed-latent supervision weight is
$\lambda_{\mathrm{comp}}=0.1$.

For training stage 3, the deterministic baseline is initialized from the best
prediction checkpoint and kept frozen. HEP is trained on a single GPU with batch
size 48 for 8k steps, using learning rate $5\times10^{-4}$, 300 warmup steps,
minimum learning-rate ratio $0.02$, and no weight decay. The rollout horizon is
also 8, and the HEP loss is averaged over all predicted steps. For the
normalized HEP cost, the denominator clamp uses $\epsilon=10^{-9}$.

\section{Details of AirGoal-10k Benchmark}
\label{sec:dataset_details}
\subsection{Dataset Construction}

We construct a UAV image-goal navigation benchmark from AirSim~\cite{airsim}, with the goal of
learning first-person visual dynamics conditioned on future motion.

\paragraph{Start points.}
Instead of uniformly sampling arbitrary poses in AirSim, which often produces
invalid viewpoints inside geometry, below the ground, or at unrealistic
altitudes, we sample start states from trajectory annotations of existing aerial VLN benchmarks~\cite{aerialvln, openfly}. This reuses poses that are already
near feasible navigation regions while still allowing us to resample new future
motions. Start states are deduplicated by scene, position, and yaw, and the held-out validation/test splits are constructed so that their start states do not overlap with the training split.

\paragraph{Future trajectory sampling.}
For each selected start state, we generate future trajectories by sampling
a target direction in the local coordinate frame of the UAV. Let the start
position and yaw be $p_0=(x_0,y_0,z_0)$ and $\psi_0$. We define the local forward,
right, and upward directions as
\begin{equation}
\begin{aligned}
e_f&=(\cos\psi_0,\sin\psi_0,0),\\
e_r&=(-\sin\psi_0,\cos\psi_0,0),\\
e_u&=(0,0,-1),
\end{aligned}
\end{equation}
where $e_u$ follows the AirSim NED convention, in which negative $z$ corresponds
to upward motion. For each trajectory, we sample
\begin{equation}
\theta \sim \mathcal{U}(0,85^\circ),\quad
\varphi \sim \mathcal{U}(0,2\pi),
\end{equation}
and construct the target direction
\begin{equation}
d_{\mathrm{target}} =
\cos\theta\, e_f
+ \sin\theta\cos\varphi\, e_r
+ \sin\theta\sin\varphi\, e_u .
\end{equation}
The angle $\theta$ controls the deviation from the initial heading, while the azimuthal angle $\varphi$ determines whether this deviation is left/right, upward/downward, or a
combined 3D turn. 

Given $d_{\mathrm{target}}$, the segment direction at each timestep is obtained
by spherical interpolation from $e_f$ to $d_{\mathrm{target}}$ with a smoothstep
schedule. The UAV position is then generated by forward integration:
\begin{equation}
\begin{aligned}
p_{i+1}&=p_i+s_i d_i,\\
s_i&=s_0(1+\epsilon_i),\\
\epsilon_i&\sim\mathcal{U}(-0.1,0.1).
\end{aligned}
\end{equation}
The base step size $s_0$ is set to match the average motion interval observed in
the existing VLN trajectories. In our implementation this corresponds to
approximately five meters per frame interval. Roll and pitch are fixed to zero,
and yaw is updated from the horizontal projection of the instantaneous motion
direction.

\paragraph{Rendering and filtering.}
AirSim is launched scene-by-scene and each accepted trajectory is rendered from
the front RGB camera. AerialVLN~\cite{aerialvln} scenes are rendered directly at $512\times512$.
OpenFly~\cite{openfly} scenes are first rendered at high resolution and then center-cropped and resized to $512\times512$. We apply automatic quality filters during collection, including invalid-height checks, black-frame rejection, and an LPIPS-based~\cite{lpips} threshold for crash detection.

\begin{figure}[t]
\centering
\includegraphics[width=\linewidth]{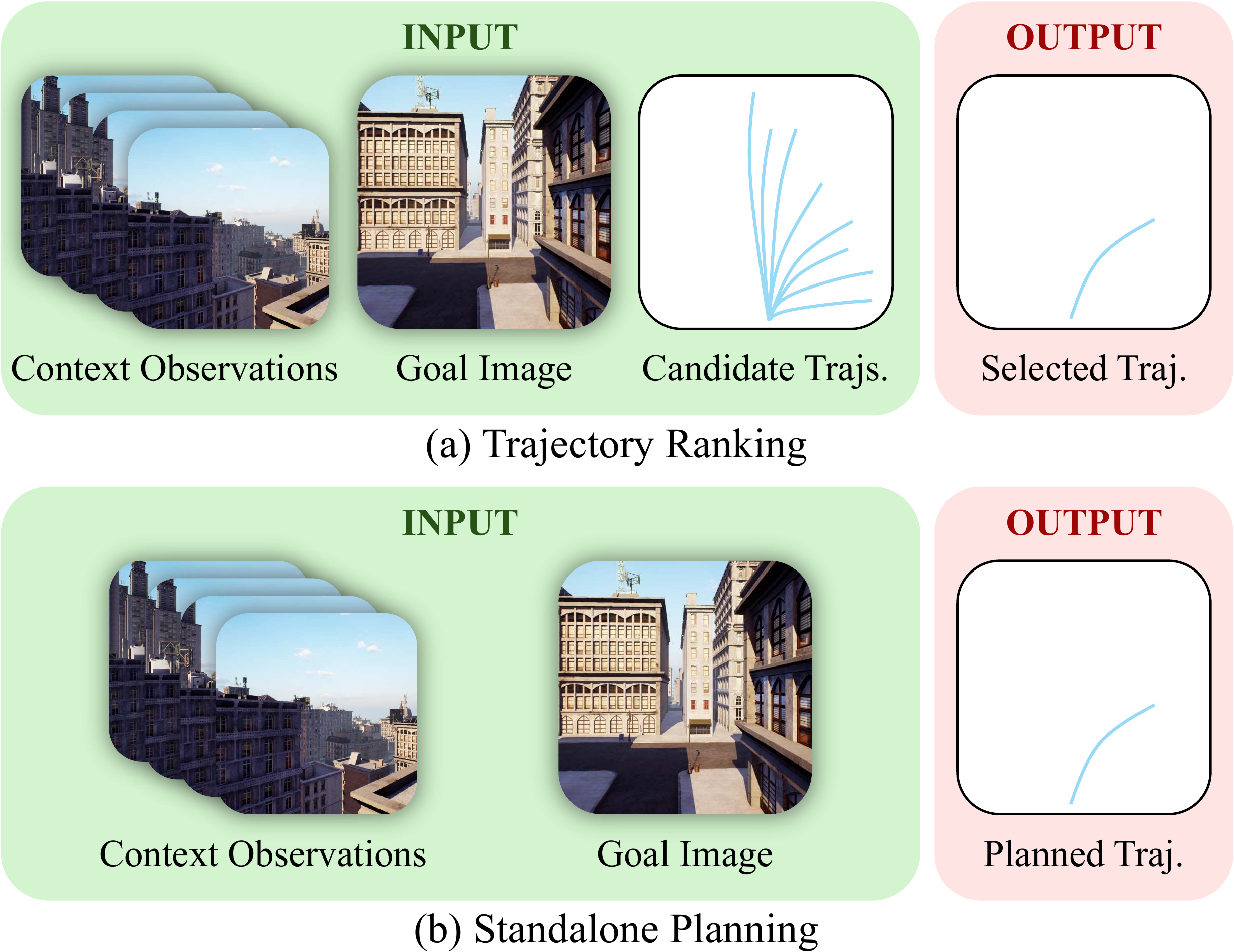}
\caption{Comparison of (a) trajectory ranking and (b) standalone planning tasks of the AirGoal-10k benchmark.}
\label{fig_task}
\end{figure}

\subsection{Task Definition}

AirGoal-10k supports two offline evaluation tasks, as illustrated in
Figure~\ref{fig_task}. Both tasks start from the same image-goal navigation
input: the agent observes a context of $C=4$ RGB frames and is given a goal image.

\paragraph{Trajectory ranking.}
In trajectory ranking, a set of candidate trajectories are provided before scoring. In our test
split, each case contains 32 externally generated candidate trajectories stored
with the trajectory metadata. These candidates share the same observed context
but branch into different 8-step future waypoint sequences. Each method scores
the same candidates for fairness and selects
\begin{equation}
    A^\star
    =
    \mathop{\mathrm{argmin}}_{A\in\mathcal{A}_{N}}
    s(A,o_g\mid O_{t-C+1:t}),
\end{equation}
where $\mathcal{A}_{N}$ denotes the first $N$ candidates and
$N\in\{8,16,32\}$. This task evaluates the quality of the trajectory scoring
function under a fixed proposal set.

\paragraph{Standalone planning.}
Standalone planning removes the external candidate set. Policy based methods can directly predict one or several trajectories, while world-model based methods use
CEM following NWM~\cite{nwm}. At each iteration, the planner samples a set of
action sequences, rolls out and scores them with the world model, keeps the
lowest-cost elites, and refits the sampling distribution over trajectory
parameters. After the last iteration, the planned trajectory is generated from
the mean parameter of the final CEM distribution. This setting tests whether
the scorer can guide search without relying on a proposal policy. In our
implementation, CEM uses 32 samples, the top 16 elites, and 3 optimization
iterations.

\paragraph{}
For both tasks, the selected or planned trajectory is compared with the
ground-truth future path using ATE and RPE. 

\begin{figure*}[t]
\centering
\includegraphics[width=\textwidth]{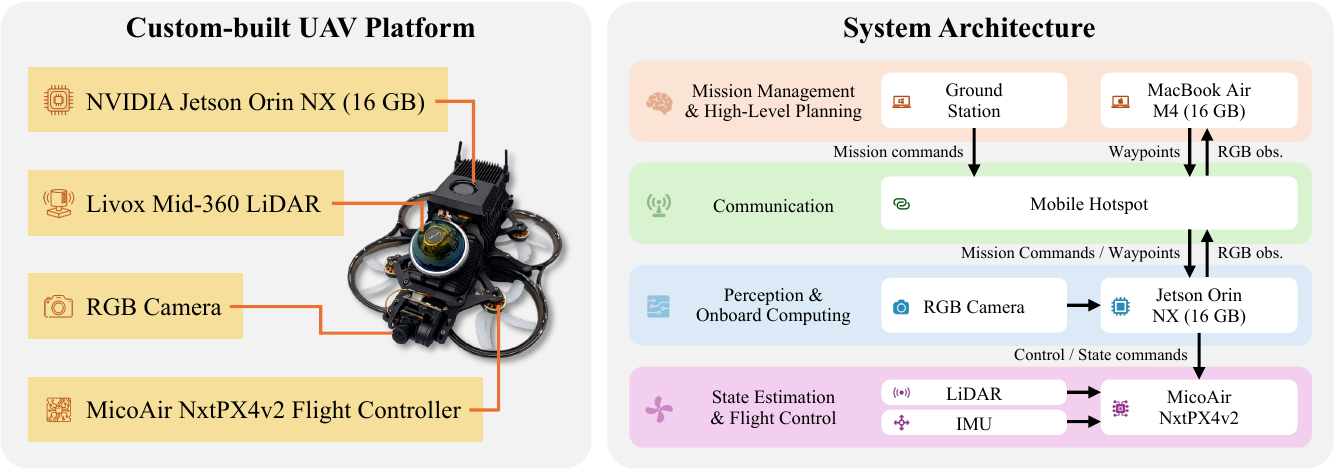}
\caption{Configuration of custom-built UAV platform and system architecture for real-world experiments.}
\label{system}
\end{figure*}

\section{Experimental Details}

\subsection{Baseline Implementation}
Since previous baselines, including policy-based~\cite{gnm,vint,nomad,flownav,navibridger} and world-model-based~\cite{nwm,onestepwm,mwm,raenwm} methods, are originally designed for 2D navigation, we modify their action encoding modules to adapt them to 3D space. For methods with different model size options, we select variants with similar parameter counts for fair comparison. For NWM~\cite{nwm} and MWM~\cite{mwm}, we adopt CDiT-B/2. For RAE-NWM~\cite{raenwm}, we adopt CDiT-B/2 and DINOv2-B~\cite{dinov2}.

All baselines are trained on the training set of AirGoal-10k until full convergence. For NaviBridger~\cite{navibridger}, we additionally train its CVAE model on AirGoal-10k in advance and use it as the learning-based prior. For One-Step WM~\cite{onestepwm}, the official implementation includes a pretraining strategy, but the corresponding pretrained checkpoint has not been released. Therefore, we train it from scratch.

During inference, policy-based methods use diffusion or flow matching to generate trajectories, and we follow their default sampling steps. NoMaD~\cite{nomad} and NaviBridger~\cite{navibridger} use 10-step diffusion, while FlowNav~\cite{flownav} uses 10-step flow matching ODE. In contrast, world models use diffusion or flow matching to generate future observations. For NWM~\cite{nwm}, the official setting with 250 sampling steps is prohibitively slow for practical applications (approximately 30 minutes per step in online simulation experiments). Therefore, following MWM~\cite{mwm}, we use 25 diffusion steps for NWM, reducing its latency to a comparable level with others. For other world models, we follow their official settings: MWM~\cite{mwm} uses 5 diffusion steps, RAE-NWM~\cite{raenwm} uses 50-step flow matching ODE, and One-Step WM~\cite{onestepwm} uses one-step shortcut flow matching.

\subsection{Offline Experiments}
All offline experiments are conducted on 4 NVIDIA RTX 4090 GPUs with 24GB memory each. All speed tests are measured on a single GPU with the same batch size. 

For the trajectory ranking task, we additionally report random selection and oracle selection strategies as references. Oracle selection always selects the trajectory with the smallest ATE with respect to the ground-truth trajectory.

\subsection{Online Simulation Experiments}
% All offline experiments are conducted on 4 NVIDIA RTX 3090 GPUs with 24GB memory each. The agent executes the first action of the best-ranked trajectory and replans after receiving the next observation. 
All online experiments are conducted on 4 NVIDIA RTX 3090 GPUs with 24GB memory each.

We evaluate online navigation on 100 start--goal pairs collected from AirSim~\cite{airsim}, including 57 AerialVLN~\cite{aerialvln} episodes and 43 OpenFly~\cite{openfly} episodes. The start-to-goal distance is computed as the three-dimensional Euclidean distance between the initial UAV position and the goal position. The average distance is 57\,m. Among the 100 episodes, 76\% have distances between 45\,m and 60\,m, while the remaining 24\% are longer-range cases between 60\,m and 80\,m.

Navigation is performed in a closed-loop manner. At each step, the UAV captures a new front-camera observation, plans its trajectory conditioned on the context observations and the goal image, and executes only the first action of the predicted trajectory. The maximum navigation budget is 20 executed actions.

% We adopt an autonomous visual stopping criterion. Goal detection is disabled for the first two UAV positions, including the initial position and the position reached after the first action, and their visual similarity scores do not update the stopping state. Starting from the third position, we compute the AlexNet LPIPS distance between the current front-camera observation and the goal image. An observation is considered a visual match when its LPIPS distance is below a threshold $\tau$, which is calibrated on an independent simulator-rendered calibration set and fixed before evaluation. A single visual match does not immediately terminate navigation. Instead, the UAV declares the goal found only when two consecutive eligible observations satisfy the LPIPS threshold. Any unmatched observation resets the consecutive-match counter.

% An episode is counted as successful only when the UAV autonomously declares the goal found and its final three-dimensional distance to the ground-truth goal is no greater than 20\,m. A visual declaration outside the 20\,m region is counted as a false positive. An episode is also considered unsuccessful if the UAV does not declare the goal found within 20 actions. Collisions and observation-capture failures are recorded as separate failure modes.
An episode is considered successful if the UAV's final three-dimensional distance to the ground-truth goal is no greater than 20\,m at termination. Observation-capture failures and collisions are recorded as separate failure modes.

We report Success Rate (SR) and Success weighted by Path Length (SPL). SPL is computed using the initial start-to-goal distance as the reference distance and the accumulated flown distance as the executed path length:
\begin{equation}
    \mathrm{SPL}
    =
    \frac{1}{N}
    \sum_{i=1}^{N}
    S_i
    \frac{L_i}
    {\max(L_i, P_i)},
\end{equation}
where $N$ is the number of episodes, $S_i \in \{0,1\}$ indicates whether episode $i$ is successful, $L_i$ denotes the initial start-to-goal distance, and $P_i$ denotes the accumulated flown distance. For runtime evaluation, we report the average planner time per online step, excluding simulator startup, environment reset, and scene-switching overhead.

\subsection{Real-world Experiments}
\label{sec:real_world_details}

To evaluate the deployability of our method in real-world environments, we develop a custom-built quadrotor platform equipped with an NVIDIA Jetson Orin NX (16\,GB), a forward-facing RGB camera, a Livox Mid-360 LiDAR, and a MicoAir NxtPX4v2 flight controller, as illustrated in Figure~\ref{system}. The RGB camera provides egocentric visual observations for image-goal navigation, while the Jetson Orin NX serves as the onboard computing and communication unit. The LiDAR measurements and onboard IMU data are fused by FAST-LIO~\cite{fastlio} to estimate the real-time position and orientation of the UAV. The resulting state estimates are provided to the low-level flight-control system, while the NxtPX4v2 flight controller is responsible for executing control commands and maintaining stable UAV motion.

The system adopts a distributed computation architecture. The UAV communicates with a Windows laptop ground station and a MacBook Air M4 (16\,GB) through a mobile-phone hotspot. The ground station is used for mission management and flight supervision, whereas world model inference and high-level navigation planning is performed locally on the MacBook. In our experiments, the MacBook runs on battery power without being plugged into an external power supply. During navigation, the current RGB observation is captured by the onboard camera and forwarded through the Jetson Orin NX to the high-level planner. Conditioned on the current observation and the goal image, UA-NWM plans with CEM, sampling 32 candidate trajectories at each of three optimization iterations, and sends the resulting waypoint command back to the UAV. The Jetson Orin NX subsequently converts the received waypoint into control commands and forwards them to the NxtPX4v2 flight controller for execution. This communication and control process is repeated in a closed-loop manner after each executed action. Navigation is terminated when the LPIPS~\cite{lpips} distance between the current observation and the goal image falls below 0.5.

Importantly, the proposed navigation model does not directly access the LiDAR measurements, IMU states, FAST-LIO pose estimates, ground-truth goal position, or ground-truth distance to the goal. FAST-LIO is used only for low-level state estimation and flight control to ensure stable and safe motion execution. High-level navigation decisions are made solely from the egocentric RGB observations and the given goal image. This separation allows us to evaluate whether the proposed method can generate executable navigation decisions on a physical UAV without relying on privileged geometric information. All real-world goal images are captured by a mobile phone, whose camera intrinsics and viewpoint characteristics differ from the onboard UAV camera. This setting further tests the robustness of UA-NWM to camera-parameter gaps between goal images and onboard observations. Additional real-world visualizations are provided in Figure~\ref{visual_real}.

\subsection{DINO Latent Visualization}
\label{sec:visualization_details}
The trajectory scoring process of UA-NWM and the deterministic baseline is performed directly in the DINOv3~\cite{dinov3} latent space. For visualizing the predicted DINOv3 latents, we adopt RAEv2~\cite{raev2} and fine-tune the officially released checkpoint on the AirGoal-10k training set to improve reconstruction quality. The decoded RGB images are only used for qualitative visualization and may not faithfully represent all information contained in the predicted DINOv3 latents. Since DINOv3 is not optimized for pixel-level reconstruction, the decoded images may exhibit blurry details or color shifts, while still providing meaningful semantic and structural cues for qualitative analysis.

The sampling procedure for other plausible states within the predicted subspace is discussed in Section~\ref{sec:further_discussion}.

\begin{table*}[t]
    \centering
    \setlength{\tabcolsep}{3pt}

    \begin{tabularx}{\textwidth}{
        >{\centering\arraybackslash}X
        *{6}{>{\centering\arraybackslash}X}
    }
        \toprule
        \multirow{2}{*}{\raisebox{-0.6ex}{Rollout Steps}}
        & \multicolumn{2}{c}{8 Candidates}
        & \multicolumn{2}{c}{16 Candidates}
        & \multicolumn{2}{c}{32 Candidates} \\

        \cmidrule(lr){2-3}
        \cmidrule(lr){4-5}
        \cmidrule(lr){6-7}

        & \mbox{ATE $\downarrow$}
        & \mbox{RPE $\downarrow$}
        & \mbox{ATE $\downarrow$}
        & \mbox{RPE $\downarrow$}
        & \mbox{ATE $\downarrow$}
        & \mbox{RPE $\downarrow$} \\
        \midrule

        1 & 1.373 & 0.378 & 1.296 & 0.358 & 1.292 & 0.357 \\
        2 & 1.319 & 0.364 & 1.244 & 0.345 & 1.185 & 0.329 \\
        3 & 1.298 & 0.358 & 1.218 & 0.337 & 1.155 & 0.322 \\
        4 & 1.301 & 0.358 & 1.197 & 0.332 & 1.134 & 0.315 \\
        5 & 1.297 & 0.357 & 1.203 & 0.333 & 1.132 & 0.314 \\
        6 & 1.286 & 0.354 & 1.198 & 0.332 & 1.114 & 0.310 \\
        7 & 1.281 & 0.354 & 1.199 & 0.332 & 1.123 & 0.312 \\
        \rowcolor{gray!20}
        8 & \textbf{1.256} & \textbf{0.347} & \textbf{1.174} & \textbf{0.326} & \textbf{1.091} & \textbf{0.304} \\

        \bottomrule
    \end{tabularx}

    \caption{Ablation study on different numbers of rollout steps
    for training.}
    \label{tab:rollout_steps_ablation}
    \vspace{15pt}

    \centering
    \setlength{\tabcolsep}{3pt}

    \begin{tabularx}{\textwidth}{
        >{\hsize=1.45\hsize
          \linewidth=\hsize
          \centering\arraybackslash}X
        >{\hsize=1.75\hsize
          \linewidth=\hsize
          \centering\arraybackslash}X
        *{6}{
            >{\hsize=0.8\hsize
              \linewidth=\hsize
              \centering\arraybackslash}X
        }
    }
        \toprule
        \multirow{2}{*}{\raisebox{-0.6ex}{Staged Training}}
        & \multirow{2}{*}{\raisebox{-0.6ex}{Delta Representation}}
        & \multicolumn{2}{c}{8 Candidates}
        & \multicolumn{2}{c}{16 Candidates}
        & \multicolumn{2}{c}{32 Candidates} \\

        \cmidrule(lr){3-4}
        \cmidrule(lr){5-6}
        \cmidrule(lr){7-8}

        & 
        & \mbox{ATE $\downarrow$}
        & \mbox{RPE $\downarrow$}
        & \mbox{ATE $\downarrow$}
        & \mbox{RPE $\downarrow$}
        & \mbox{ATE $\downarrow$}
        & \mbox{RPE $\downarrow$} \\
        \midrule

        $\times$ & $\checkmark$
        & 1.309 & 0.360
        & 1.222 & 0.338
        & 1.189 & 0.330 \\

        $\checkmark$ & $\times$
        & 1.370 & 0.375
        & 1.309 & 0.359
        & 1.263 & 0.347 \\

        \rowcolor{gray!20}
        $\checkmark$ & $\checkmark$
        & \textbf{1.256} & \textbf{0.347}
        & \textbf{1.174} & \textbf{0.326}
        & \textbf{1.091} & \textbf{0.304} \\

        \bottomrule
    \end{tabularx}

    \caption{Ablation study on the staged training strategy and delta representation~\cite{deltatok}}
    \label{tab:staged_delta_ablation}
    \vspace{15pt}

    \centering
    \setlength{\tabcolsep}{3pt}

    \begin{tabularx}{\textwidth}{
        *{2}{
            >{\hsize=0.75\hsize
              \linewidth=\hsize
              \centering\arraybackslash}X
        }
        *{6}{
            >{\hsize=1.0833\hsize
              \linewidth=\hsize
              \centering\arraybackslash}X
        }
    }
        \toprule
        \multirow{2}{*}{\raisebox{-0.6ex}{$K$}}
        & \multirow{2}{*}{\raisebox{-0.6ex}{$M$}}
        & \multicolumn{2}{c}{8 Candidates}
        & \multicolumn{2}{c}{16 Candidates}
        & \multicolumn{2}{c}{32 Candidates} \\

        \cmidrule(lr){3-4}
        \cmidrule(lr){5-6}
        \cmidrule(lr){7-8}

        &
        & \mbox{ATE $\downarrow$}
        & \mbox{RPE $\downarrow$}
        & \mbox{ATE $\downarrow$}
        & \mbox{RPE $\downarrow$}
        & \mbox{ATE $\downarrow$}
        & \mbox{RPE $\downarrow$} \\
        \midrule

        32 & 16
        & 1.342 & 0.370
        & 1.264 & 0.350
        & 1.204 & 0.334 \\

        32 & 64
        & 1.359 & 0.374
        & 1.289 & 0.355
        & 1.233 & 0.341 \\

        16 & 32
        & 1.309 & 0.360
        & 1.235 & 0.341
        & 1.159 & 0.321 \\

        64 & 32
        & 1.309 & 0.360
        & 1.230 & 0.340
        & 1.155 & 0.320 \\

        \rowcolor{gray!20}
        32 & 32
        & \textbf{1.256} & \textbf{0.347}
        & \textbf{1.174} & \textbf{0.326}
        & \textbf{1.091} & \textbf{0.304} \\

        \bottomrule
    \end{tabularx}

    \caption{Ablation study on different combinations of $K$ and $M$.}
    \label{tab:km_ablation}
    \vspace{15pt}

    \centering
    \setlength{\tabcolsep}{5pt}

    \begin{tabularx}{\textwidth}{l*{4}{>{\centering\arraybackslash}X}}
        \toprule
        \multirow{2}{*}{\raisebox{-0.6ex}{Methods}}
        & \multicolumn{2}{c}{RECON}
        & \multicolumn{2}{c}{GO Stanford} \\

        \cmidrule(lr){2-3}
        \cmidrule(lr){4-5}

        & \mbox{ATE $\downarrow$}
        & \mbox{RPE $\downarrow$}
        & \mbox{ATE $\downarrow$}
        & \mbox{RPE $\downarrow$} \\
        \midrule

        NWM~\cite{nwm} + NoMaD ($\times 8$)
        & 1.01 & 0.27 & 2.06 & 0.59 \\
         
        NWM~\cite{nwm} + NoMaD ($\times 16$)
        & 1.00 & 0.27 & 2.11 & 0.60 \\
         
        NWM~\cite{nwm} + NoMaD ($\times 32$)
        & 0.99 & 0.27 & 2.11 & 0.60 \\
         
        \midrule

        RAE-NWM~\cite{raenwm} + NoMaD ($\times 8$)
        & 1.07 & 0.30 & 1.97 & 0.57 \\
         
        RAE-NWM~\cite{raenwm} + NoMaD ($\times 16$)
        & 1.09 & 0.31 & 2.02 & 0.58 \\
         
        RAE-NWM~\cite{raenwm} + NoMaD ($\times 32$)
        & 1.11 & 0.32 & 2.01 & 0.59 \\

        % Deterministic Baseline (Ours) + NoMaD ($\times 8$)
        % & 1.443 & 0.394 & 1.925 & 0.556 \\

        % Deterministic Baseline (Ours) + NoMaD ($\times 16$)
        % & 1.616 & 0.437 & 1.920 & 0.553 \\

        % Deterministic Baseline (Ours) + NoMaD ($\times 32$)
        % & 1.716 & 0.464 & 2.002 & 0.575 \\

        \midrule

        \rowcolor{gray!20}
        UA-NWM (Ours) + NoMaD ($\times 8$)
        & 0.94 & \textbf{0.25} & 1.68 & 0.49 \\

        \rowcolor{gray!20}
        UA-NWM (Ours) + NoMaD ($\times 16$)
        & \textbf{0.92} & \textbf{0.25} & 1.52 & \textbf{0.44} \\

        \rowcolor{gray!20}
        UA-NWM (Ours) + NoMaD ($\times 32$)
        & \textbf{0.92} & \textbf{0.25} & \textbf{1.51} & \textbf{0.44} \\

        \bottomrule
    \end{tabularx}

    \caption{Comparison of NWM~\cite{nwm},RAE-NWM~\cite{raenwm} and UA-NWM
    on the RECON~\cite{recon} and GO Stanford~\cite{gostanford} datasets using different numbers
    of trajectories generated by NoMaD~\cite{nomad}.}
    \label{tab:recon_gostanford}
\end{table*}

\section{Additional Ablation Studies}
\subsection{Rollout Steps for Training}
As described in Section~\ref{sec:training_inference_procedure}, we employ autoregressive rollout in stages 2 and 3 and average the losses across all rollout steps to mitigate error accumulation in autoregressive world models. We vary the number of rollout steps to investigate its effect on navigation performance. As shown in Table~\ref{tab:rollout_steps_ablation}, performance generally improves with longer rollouts. Exposing the model to its own intermediate predictions during training better matches autoregressive inference and improves robustness to accumulated errors. We therefore use an eight-step rollout in both Stages 2 and 3.

\subsection{Staged Training and Delta Representation}
As described in Section~\ref{sec:training_inference_procedure}, our deterministic backbone adopts a staged training strategy that first learns compact representations and latent transitions before learning future prediction. Following DeltaWorld~\cite{deltatok}, we further employ a delta representation to encode the change in visual representations between adjacent frames, allowing the predictor to estimate this delta rather than the full representation of the next frame. We ablate both designs in Table~\ref{tab:staged_delta_ablation}.

Training the deterministic backbone end-to-end without staged training degrades performance, indicating that pretraining the compressor, fuser, and delta decoder to model compact representations and latent transitions provides a more stable and effective initialization for future prediction. Replacing delta prediction with direct prediction of the full next-frame representation causes a larger performance drop across different numbers of candidate trajectories. This suggests that predicting inter-frame changes reduces the redundancy shared by adjacent observations and enables the predictor to focus on temporally varying information, thereby facilitating more accurate dynamics modeling. Combining both strategies yields the best overall performance.

\subsection{Number of Compressed and Delta Tokens}
We further investigate the effects of $K$ and $M$ by evaluating different combinations of the two hyperparameters. As shown in Table~\ref{tab:km_ablation}, the setting $K=32$ and $M=32$ achieves the best performance across all candidate-set sizes. Deviating from this configuration in either direction consistently leads to inferior results, indicating that both representations benefit from a moderate capacity. Smaller values may limit their expressive power, whereas larger values may introduce unnecessary redundancy and make optimization more difficult. We therefore adopt $K=32$ and $M=32$ as the default setting.

\section{Performance on 2D Navigation Benchmarks}

We further evaluate UA-NWM on two 2D ground image-goal navigation benchmarks, RECON~\cite{recon} and GO Stanford~\cite{gostanford}, and compare it with NWM~\cite{nwm} and RAE-NWM~\cite{raenwm}. We use the official inference settings of 250 diffusion steps for NWM and 50 flow-matching ODE steps for RAE-NWM. For fair comparison, all methods rank the same candidate trajectory sets generated by NoMaD~\cite{nomad}. As shown in Table~\ref{tab:recon_gostanford}, UA-NWM consistently outperforms both baselines across the two datasets and all candidate-set sizes. Although UA-NWM is primarily designed for aerial navigation in 3D environments, these results demonstrate that its uncertainty-aware trajectory scoring mechanism also transfers effectively to 2D ground navigation, indicating broader applicability beyond the original setting.

% The discrepancy between the NWM results on RECON and those reported in the original paper is likely due to differences in the candidate trajectory sets, as the exact candidate sets used in the NWM paper are not publicly available.

\section{Further Analysis of HEP}

This section provides a closer look at the proposed Hierarchical Error
Projection (HEP). We use a probabilistic view to clarify why the
unexplained residual can serve as a trajectory-goal compatibility score, and
then describe what the coarse-to-fine projection actually computes. We also
discuss how the training objective encourages the predicted bases to capture
residual directions that recur under similar context-action conditions. Our
purpose is to make the assumptions behind the HEP interpretation explicit,
rather than to claim that HEP estimates a fully calibrated likelihood of future
observations.

\subsection{Residual Energy as a Conditional Compatibility Score}

For a fixed observation context $O_{t-C+1:t}$ and a candidate action sequence
$A_{t:t+H-1}$, let $\mu$ be the deterministic DINO feature prediction and
$x_g=\phi(o_g)$ be the DINO feature of the goal image. UA-NWM computes the discrepancy
\begin{equation}
    e=\mathrm{norm}(x_g)-\mathrm{norm}(\mu).
\end{equation}
Let $E$ denote the residual induced by plausible futures under the same context and action. Suppose that most plausible
variation around $\mu$ lies in an $R$-dimensional subspace
$\mathcal{S}=\mathrm{span}(V)$, where the columns of $V$ are orthonormal. A
simple conditional residual model is
\begin{equation}
    E = V a + \xi,
    \quad
    a\sim \mathcal{N}(0,\tau^2 I),
    \quad
    \xi\sim \mathcal{N}(0,\sigma^2 I).
\end{equation}
Here, $Va$ denotes uncertainty-explainable variation and $\xi$ denotes residual
variation outside the plausible subspace. The covariance is
\begin{equation}
    \Sigma = \tau^2 VV^\top+\sigma^2 I .
\end{equation}

For this Gaussian residual model, the negative log-likelihood of an observed
residual contains the quadratic Mahalanobis term $e^\top\Sigma^{-1}e$; with
fixed $R$, $\tau$, and $\sigma$, the remaining log-determinant term is constant
with respect to the observed residual. Decomposing any residual as
$e=e^\parallel+e^\perp$, with $e^\parallel\in\mathcal{S}$ and
$e^\perp\perp\mathcal{S}$, gives
\begin{equation}
    e^\top\Sigma^{-1}e
    =
    \frac{\|e^\parallel\|_2^2}{\tau^2+\sigma^2}
    +
    \frac{\|e^\perp\|_2^2}{\sigma^2}.
\end{equation}
The exact Gaussian score therefore penalizes both components, but with different
weights. UA-NWM uses only the second term, corresponding to the
high-anisotropy regime where plausible variation inside $\mathcal{S}$ is much
larger than unexplained variation outside it, i.e., $\tau^2\gg\sigma^2$. In this
case, the penalty on $e^\parallel$ is strongly down-weighted and the dominant
criterion becomes $\|e^\perp\|_2^2$. If $\tau^2$ is not substantially larger than
$\sigma^2$, a calibrated quadratic score would also retain a non-negligible
penalty on $e^\parallel$; our score should then be viewed as a conservative
compatibility approximation that treats variation inside $\mathcal{S}$ as
acceptable for navigation. This yields the OOD interpretation in the main paper: a candidate is penalized primarily when the goal discrepancy cannot be
explained by the predicted uncertainty subspace, rather than simply when it differs from the deterministic mean prediction.

\subsection{What the Hierarchical Projection Computes}

The full feature discrepancy is a spatial field over $N$ DINO patches. HEP
constructs a multi-scale approximation of the plausible residual subspace using
grid scales $\mathcal{G}=\{1,2,7,14\}$. At scale $g$, the patch grid is partitioned
into cells $\Omega_{g,c}$. Let $r_{g,i}$ denote the residual at patch $i$ when
entering scale $g$, initialized by $r_{1,i}=e_i$. The cell-average residual is
\begin{equation}
    \bar{r}_{g,c}
    =
    \frac{1}{|\Omega_{g,c}|}
    \sum_{i\in\Omega_{g,c}} r_{g,i}.
\end{equation}
The bar indicates that this is an average over all patches in the cell, not the
residual of a single patch. For each cell, HEP predicts a basis
$U_{g,c}=\{u_{g,c,1},\ldots,u_{g,c,R}\}$. It then computes a ridge-regularized
projection of $\bar r_{g,c}$ onto the span of this basis:
\begin{equation}
    \alpha_{g,c}
    =
    \mathop{\mathrm{argmin}}_{\alpha\in\mathbf{R}^{R}}
    \left\|
    \bar{r}_{g,c}
    -
    \sum_{r=1}^{R}\alpha_r u_{g,c,r}
    \right\|_2^2
    +
    \lambda\|\alpha\|_2^2 ,
\end{equation}
\begin{equation}
    p_{g,c}=\sum_{r=1}^{R}\alpha_{g,c,r}u_{g,c,r}.
\end{equation}
When $\lambda=0$, $p_{g,c}$ is the ordinary least-squares projection of
$\bar r_{g,c}$ onto $\mathrm{span}(U_{g,c})$. With $\lambda>0$, it is a
regularized projection that improves numerical stability and prevents unstable coefficients
when the predicted basis vectors are correlated. The projected vector is then
subtracted from every patch in the cell:
\begin{equation}
    r_{g^+,i}=r_{g,i}-p_{g,c},
    \quad i\in\Omega_{g,c},
\end{equation}
where $g^+$ denotes the next finer scale.

This operation has a clear geometric meaning. At each scale, HEP removes a
low-rank, cell-wise constant residual component. Coarse scales explain broad
spatially coherent deviations, while fine scales explain more localized
deviations. The finest $14\times14$ scale reduces to patch-wise residual
projection because each cell contains one DINO patch.

We distinguish HEP from a single exact orthogonal projection. Let
$\mathcal{S}_g$ denote the space of residual fields that are constant within
each cell of scale $g$ and whose cell value lies in the corresponding local
basis span, and let
\begin{equation}
    \mathcal{S}_{\mathrm{all}}
    =
    \mathcal{S}_1+\mathcal{S}_2+\mathcal{S}_7+\mathcal{S}_{14}.
\end{equation}
An exact orthogonal projection would solve one global least-squares problem,
\begin{equation}
    p^\star
    =
    \mathop{\mathrm{argmin}}_{p\in\mathcal{S}_{\mathrm{all}}}
    \|e-p\|_2^2 ,
\end{equation}
and would produce a residual $e-p^\star$ that is orthogonal to the whole space
$\mathcal{S}_{\mathrm{all}}$. This global solve is not what HEP implements.

Instead, HEP performs a one-pass coarse-to-fine residual decomposition. It first
removes the component predicted at the coarsest scale, then applies the next
scale to the remaining residual, and continues until the finest scale. This is
greedy because each scale explains only the residual left by previous scales and
the earlier components are not jointly re-optimized. If the scale-wise subspaces
are mutually orthogonal, this sequential procedure is equivalent to the exact
orthogonal projection onto their direct sum, since components from different
scales do not interfere with each other. The implementation does not explicitly
enforce this condition. Therefore, in the general case, the final residual should
be interpreted as the unexplained component left by the hierarchical projection
procedure, rather than as the exact Euclidean distance to a globally projected
subspace. This residual is the quantity used by UA-NWM for trajectory scoring.

\subsection{Why the Training Objective Learns Plausible Residual Directions}

During HEP training, the deterministic baseline is fixed and HEP is optimized
with the normalized residual loss
\begin{equation}
    \mathcal{L}_{\mathrm{HEP}}
    =
    \frac{
    \sum_{i=1}^{N}\|e_{t+1,i}^{\perp}\|_2^2
    }{
    \sum_{i=1}^{N}\|e_{t+1,i}\|_2^2+\epsilon
    } .
\end{equation}
This objective encourages the predicted bases to explain residual directions
that repeatedly occur under the same type of context and action. To connect the
loss to subspace learning, consider a fixed scale-cell pair $(g,c)$ and
temporarily view its predicted basis span as a fixed $R$-dimensional subspace
$\mathcal{S}$. For a single training sample, the cell-average residual
$\bar r_{g,c}$ is decomposed into an explained part
$P_{\mathcal{S}}\bar r_{g,c}$ and an unexplained part
$(I-P_{\mathcal{S}})\bar r_{g,c}$. The loss reduces this unexplained
component, while the denominator in $\mathcal{L}_{\mathrm{HEP}}$ only rescales
the sample contribution.

Over the whole training set, minimizing this residual term corresponds to
minimizing the empirical average of the unexplained energy. In population form,
this empirical average is written as
\begin{equation}
    \mathbb{E}\left[
    \|(I-P_{\mathcal{S}})\bar{\rho}_{g,c}\|_2^2
    \right],
\end{equation}
where $\bar{\rho}_{g,c}$ denotes the random variable corresponding to the
cell-average residual $\bar r_{g,c}$ when a transition is sampled from the
training distribution, and $P_{\mathcal{S}}$ is the projection onto
$\mathcal{S}$. Minimizing this quantity is equivalent to maximizing the expected
explained energy
\begin{equation}
    \mathbb{E}\left[
    \|P_{\mathcal{S}}\bar{\rho}_{g,c}\|_2^2
    \right].
\end{equation}
Thus, for a fixed context-action condition, the optimal low-rank subspace
contains the principal residual directions of plausible futures. HEP does not
store a separate subspace for each condition. Instead, it learns a conditional
mapping from the deterministic prediction and context features to the local
basis $U_{g,c}$. Since the deterministic prediction $\mu$ depends on the
candidate action sequence, different candidate trajectories can induce
different predicted bases and therefore different uncertainty directions.

This argument also clarifies why the design does not simply memorize arbitrary
training residuals. Each cell predicts only $R$ basis directions, so it cannot
explain all channel-space errors when $R\ll D$, where $D$ is the DINO feature-channel dimension. At coarser scales, the same
projected vector is subtracted from all patches in a cell, so only spatially
coherent deviations can be explained at that scale. In addition, the
deterministic baseline is frozen during HEP training, which prevents HEP from
changing the mean prediction $\mu$ to absorb residual errors. These restrictions
make the easiest residuals to explain those that appear consistently under
similar context-action conditions. Therefore, the HEP loss provides a
principled surrogate for learning conditional plausible-variation directions,
while the final navigation score uses the remaining unexplained residual to
distinguish goal-compatible and goal-incompatible candidate trajectories.

\begin{figure}[t]
\centering
\includegraphics[width=\linewidth]{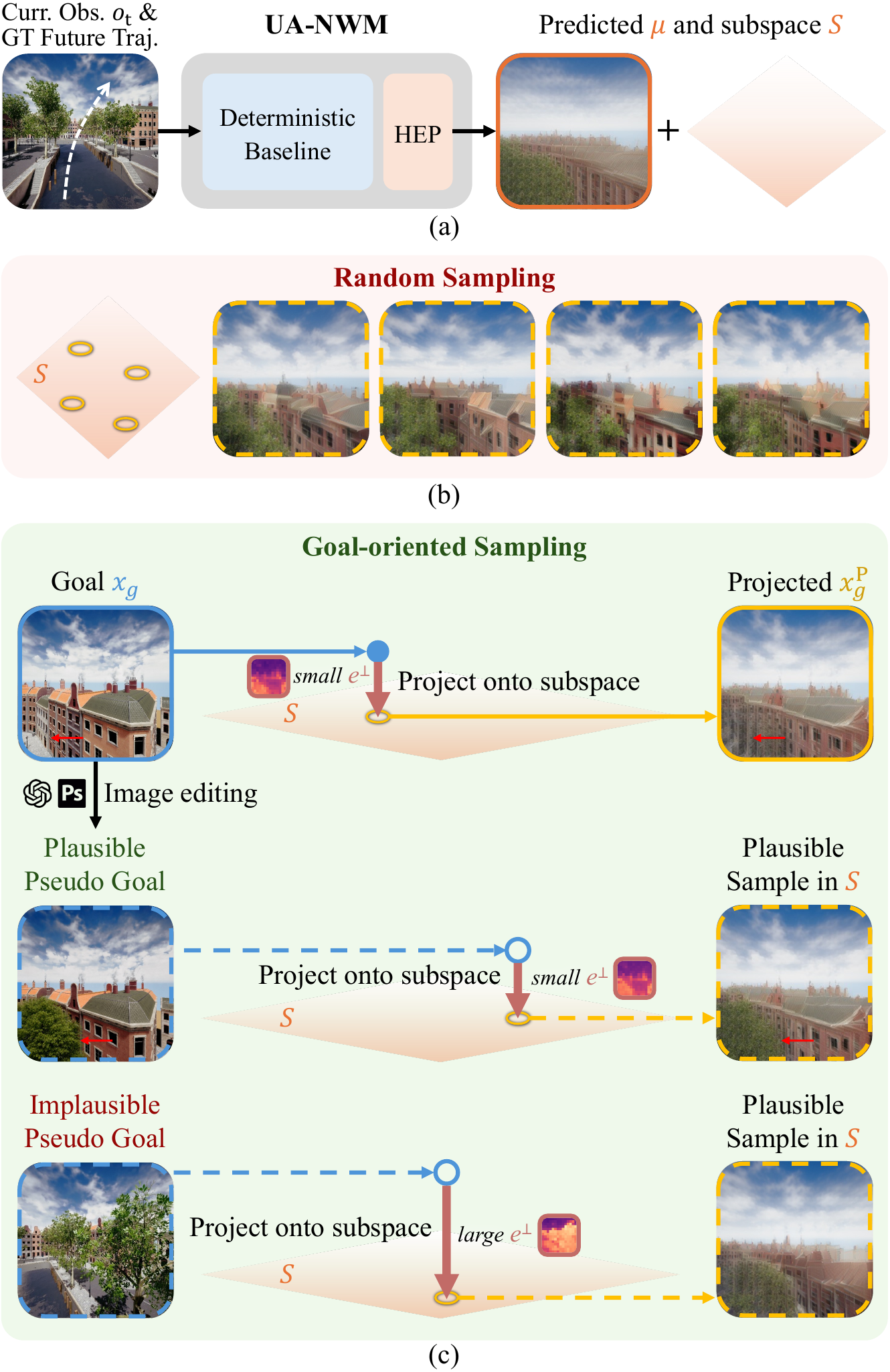}
\caption{Different strategies of sampling plausible states on predicted subspace.}
\label{fig_sample}
\end{figure}

\section{Further Discussions and Future Work}
\label{sec:further_discussion}

In Figure~\ref{fig_visual_compare}, we visualize alternative plausible states sampled from the predicted uncertainty subspace. Here, we detail the sampling procedure and further discuss the capability and role of UA-NWM as a world model.

As illustrated in Figure~\ref{fig_sample}(a), UA-NWM first predicts the deterministic future state $\mu$ and the uncertainty subspace $\mathcal{S}$. We then sample different points within $\mathcal{S}$ to obtain $x_g^P$ and other plausible future states.

The state $x_g^P$ can be interpreted as the point in $\mathcal{S}$ that is closest to the goal state $x_g$. It is obtained as $\mu+e^{\parallel}$, which can be viewed as projecting $x_g$ onto $\mathcal{S}$. Compared with the discrepancy between $\mu$ and $x_g$, the decoded $x_g^P$ typically exhibits a substantially smaller gap from $x_g$. It also tends to preserve a plausible appearance because it is derived from $\mu$ through hierarchical error projection, which is trained to explain residuals along plausible uncertainty directions.

Nevertheless, although HEP is trained to capture plausible residual directions, $\mathcal{S}$ remains only a low-dimensional, local approximation to the conditional future-state distribution. Consequently, not every point in $\mathcal{S}$ lies in a high-density region of the true distribution, and random sampling in it does not always yield high-quality results. As shown in Figure~\ref{fig_sample}(b), these samples exhibit meaningful diversity from the mean prediction $\mu$ and generally preserve reasonable global layouts, but their spatial coherence is limited. In particular, different regions of the same building may exhibit inconsistent colors, making the image resemble a shuffled jigsaw puzzle. This likely arises because, although the hierarchical projection captures some spatial correlations across image regions, the resulting uncertainty subspace remains insufficient to enforce fully coherent variations among correlated patches. When multiple building colors are plausible, independent sampling within $\mathcal{S}$ may assign different colors to correlated patches rather than selecting a coherent appearance for the entire structure.

To obtain more interpretable samples, we introduce a goal-oriented sampling strategy. Inspired by the construction of $x_g^P$, we generate alternative plausible samples from modified goal images. As shown in Figure~\ref{fig_sample}(c), we first use image-editing tools to construct a visually distinct yet contextually plausible pseudo-goal image from the original goal image $x_g$. If the pseudo-goal is plausible under the given initial observation and action sequence, it should lie near a high-density region of the same conditional future-state distribution. Projecting it onto $\mathcal{S}$ using HEP should therefore produce a plausible state that remains close to the pseudo-goal while differing in appearance from $x_g^P$.

Conversely, if the pseudo-goal is incompatible with the initial observation or action sequence, its projection will not closely reproduce the pseudo-goal, but will instead remain constrained by the plausible future states represented by $\mathcal{S}$. In the last example of Figure~\ref{fig_sample}(c), the pseudo-goal is captured from another 3D position in the simulation environment and corresponds to the endpoint of a different trajectory. It is therefore incompatible with the conditional future-state distribution induced by the original trajectory. After projection onto the original subspace $\mathcal{S}$, the resulting state preserves the overall layout associated with the original trajectory rather than reproducing the pseudo-goal. The resulting large projection discrepancy allows UA-NWM to identify the pseudo-goal as incompatible with the predicted future-state distribution.

These observations suggest that UA-NWM is effective as a navigation discriminator and also shows promise as an efficient stochastic generator. Through hierarchical error projection, it can assess whether a candidate trajectory is compatible with the goal image while accounting for future-state uncertainty, making it well suited to image-goal navigation. However, fully realizing future-distribution prediction for efficient stochastic generation requires a more expressive uncertainty representation. Future work may explore more expressive uncertainty representations that capture spatial coherence across patches and temporal coherence across frames more precisely, enabling stochastic future generation by predicting a future-state distribution in a single forward pass and sampling diverse futures from it without running the full model again for each sample.

\section{Additional Visualization Results}
\label{sec:additional_visualization}
\subsection{Hierarchical Error Projection}
\begin{figure*}[t]
\centering
\includegraphics[width=0.84\textwidth]{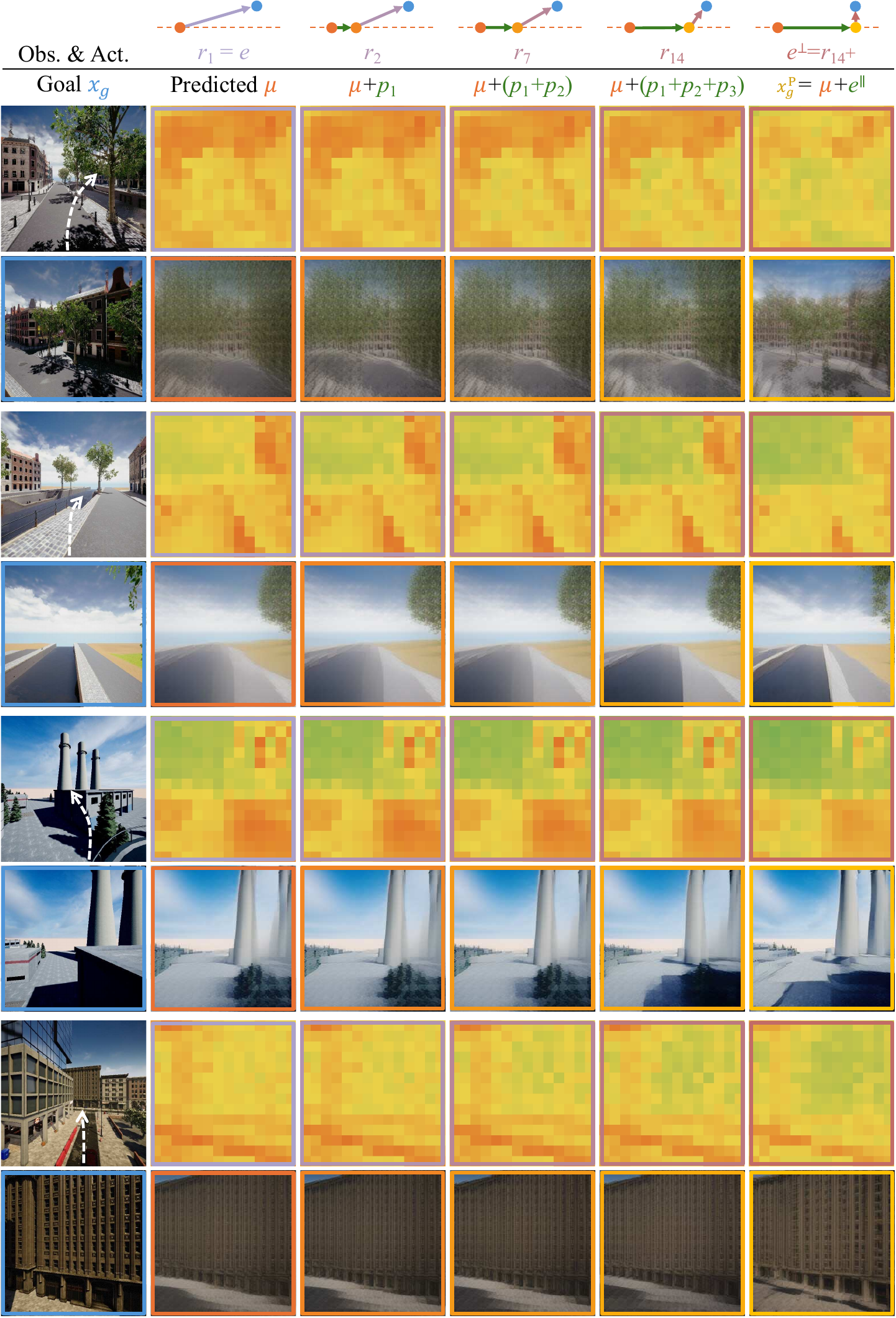}
\caption{Visualizations of hierarchical error projection process.}
\label{visual_hep}
\end{figure*}

Visualizations of the hierarchical error projection process are shown in Figure~\ref{visual_hep}. As the projection proceeds, the unexplained residual progressively decreases. Starting from $\mu$, we gradually add the explained component $e^{\parallel}$ on it, ultimately obtaining $x_g^P$ that closely resembles $x_g$.

\subsection{Uncertainty Subspace}
\begin{figure*}[t]
\centering
\includegraphics[width=\textwidth]{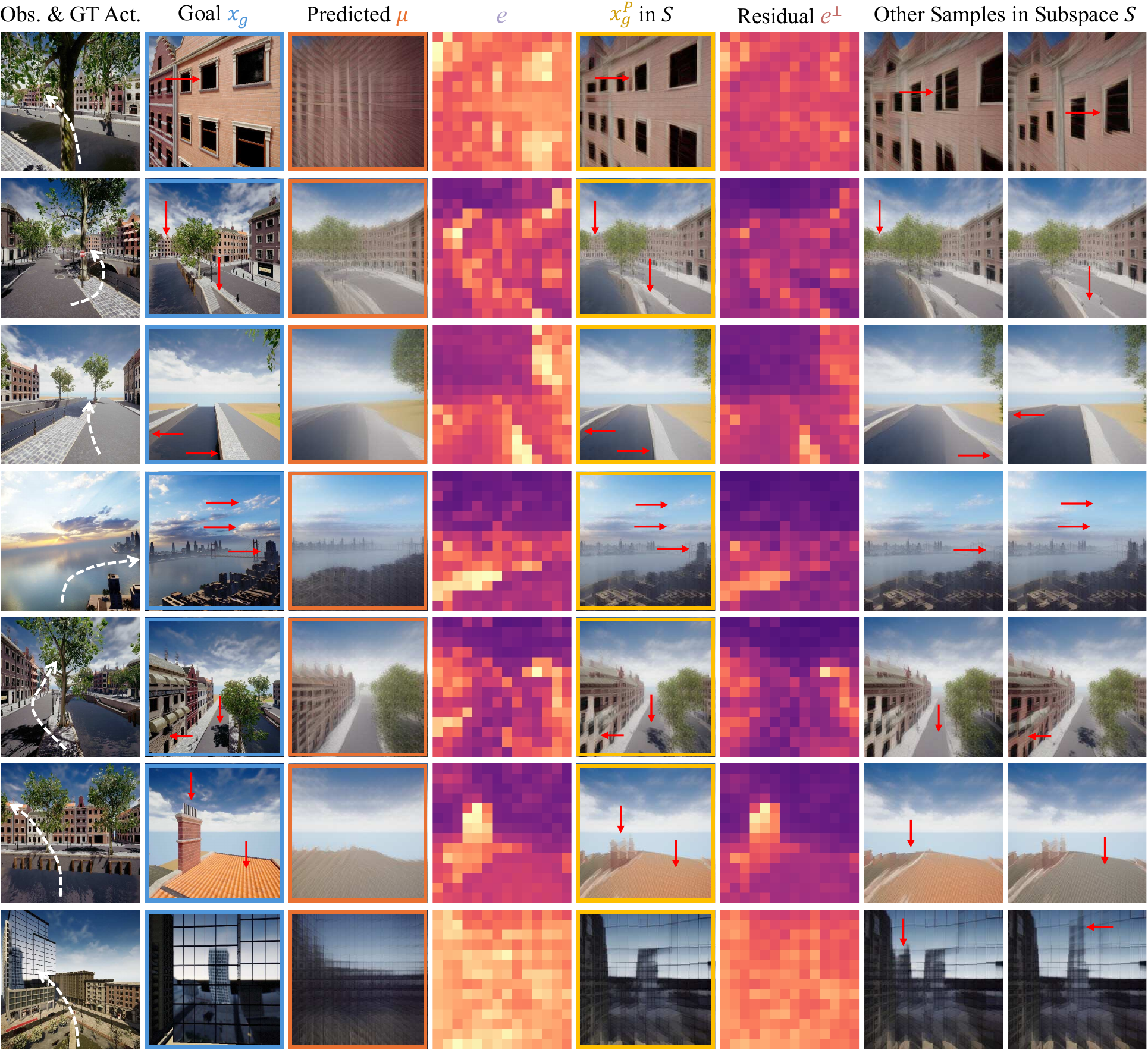}
\caption{Additional visualizations of samples from the predicted uncertainty subspace.}
\label{visual_uncertainty}
\end{figure*}

Additional visualizations of samples from the predicted uncertainty subspace are shown in Figure~\ref{visual_uncertainty}. For each case, besides $x_g^P$, we identify two alternative plausible samples in $\mathcal{S}$, demonstrating that the predicted subspace captures meaningful future-state diversity, primarily arising from occlusion-induced ambiguity or long-horizon drift.

\subsection{Offline Trajectory Ranking}
\begin{figure*}[t]
\centering
\includegraphics[width=\textwidth]{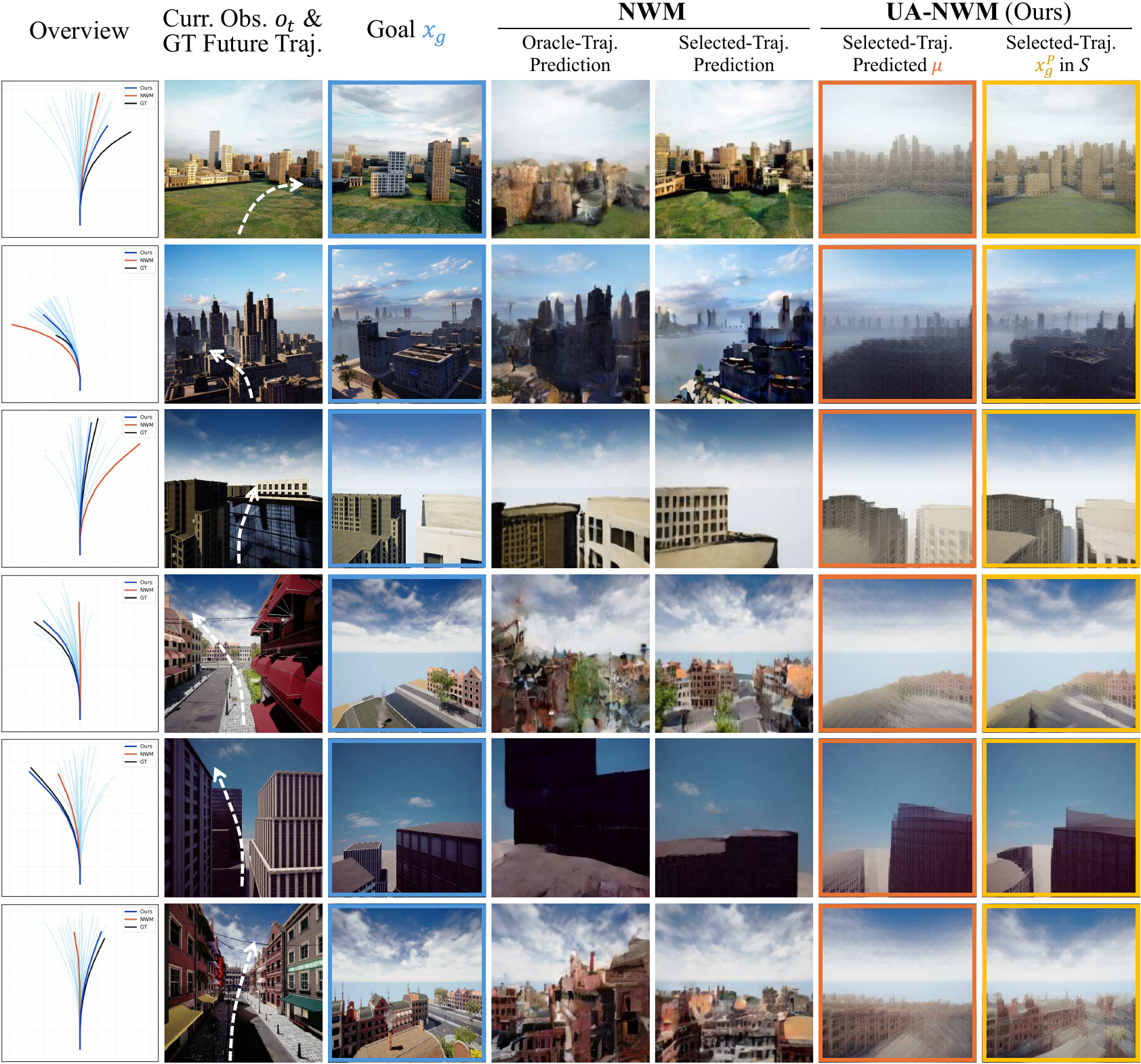}
\caption{Qualitative comparisons on the trajectory ranking task.}
\label{visual_ranking}
\end{figure*}

Qualitative comparisons on the offline trajectory ranking task are shown in Figure~\ref{visual_ranking}. When ranking 32 candidates, NWM~\cite{nwm} is susceptible to error accumulation and sampling variability, which may cause it to favor inferior trajectories. In contrast, UA-NWM leverages HEP to score trajectories using the unexplained residual, yielding more robust rankings that favor candidates closer to the ground-truth trajectory under future-state uncertainty.

\subsection{Offline Standalone Planning}
\begin{figure*}[t]
\centering
\includegraphics[width=\textwidth]{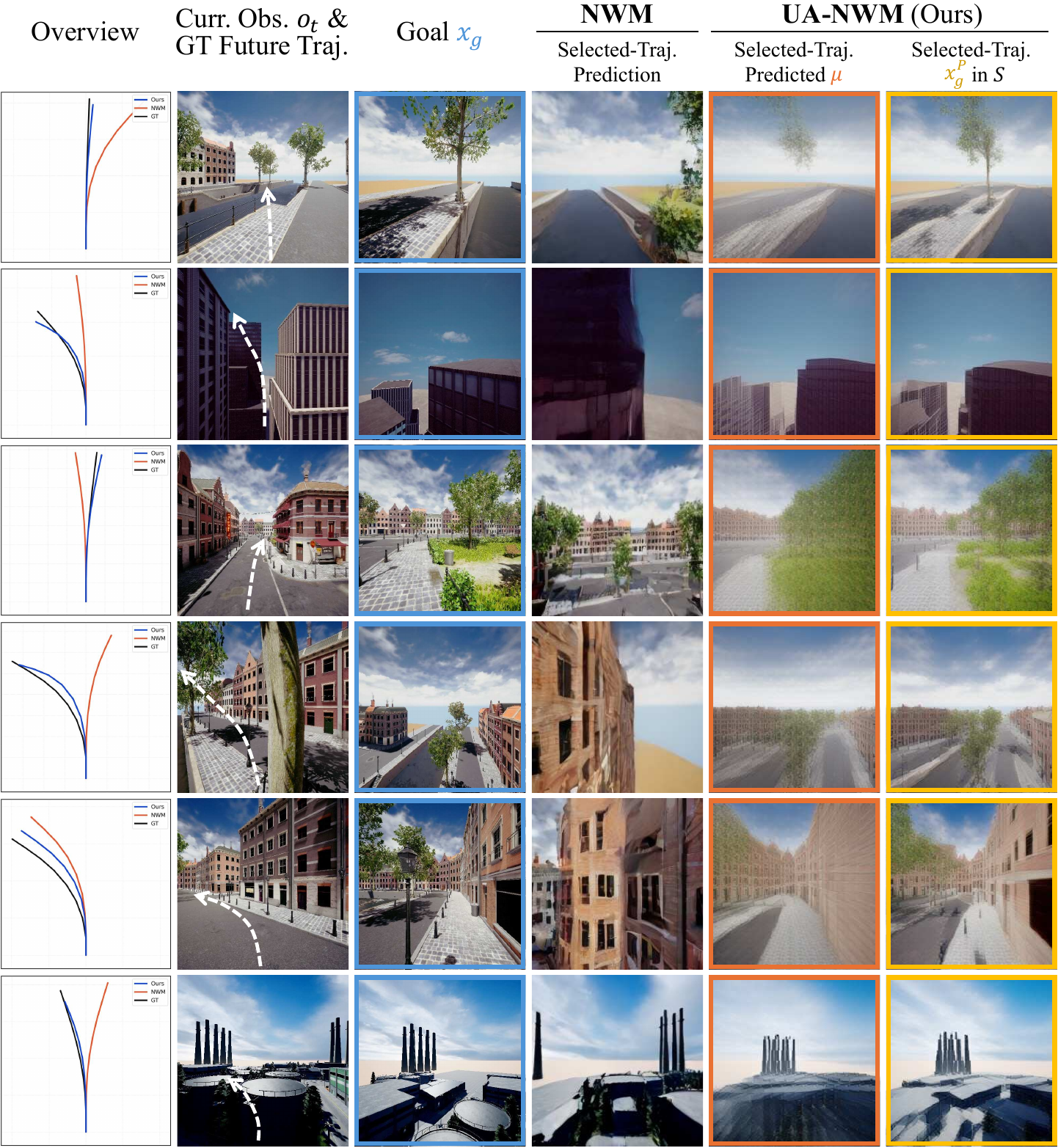}
\caption{Qualitative comparisons on the standalone planning task.}
\label{visual_planning1}
\end{figure*}

\begin{figure*}[t]
\centering
\includegraphics[width=\textwidth]{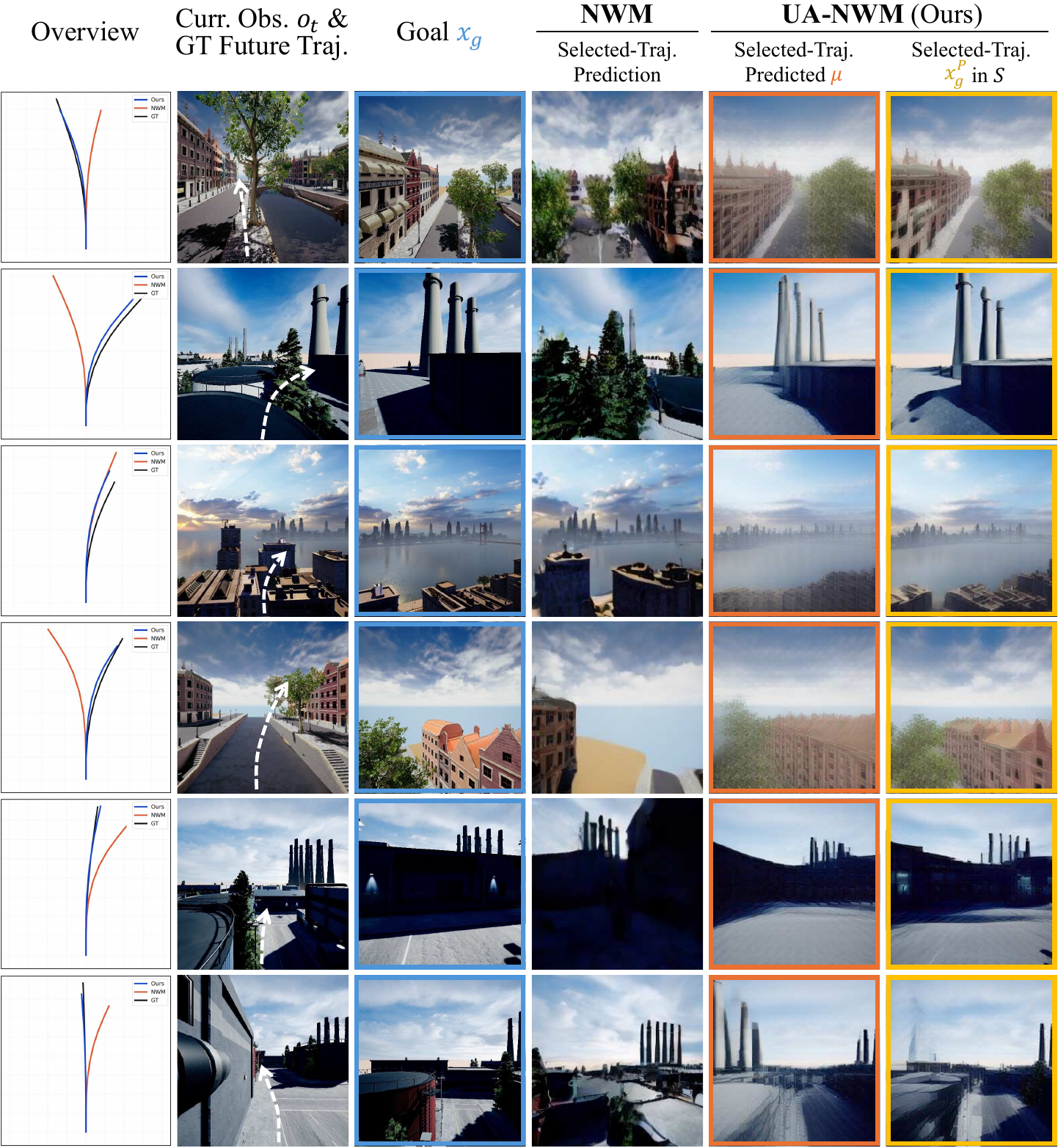}
\caption{Qualitative comparisons on the standalone planning task.}
\label{visual_planning2}
\end{figure*}

Qualitative comparisons on the offline standalone planning task are shown in Figures~\ref{visual_planning1} and~\ref{visual_planning2}. Under the same CEM planning framework, UA-NWM effectively identifies the direction toward the goal through iterative sampling, prediction, and optimization, whereas NWM~\cite{nwm} often fails to recover the correct path.

\subsection{Online Simulation}
\begin{figure*}[t]
\centering
\includegraphics[width=\textwidth]{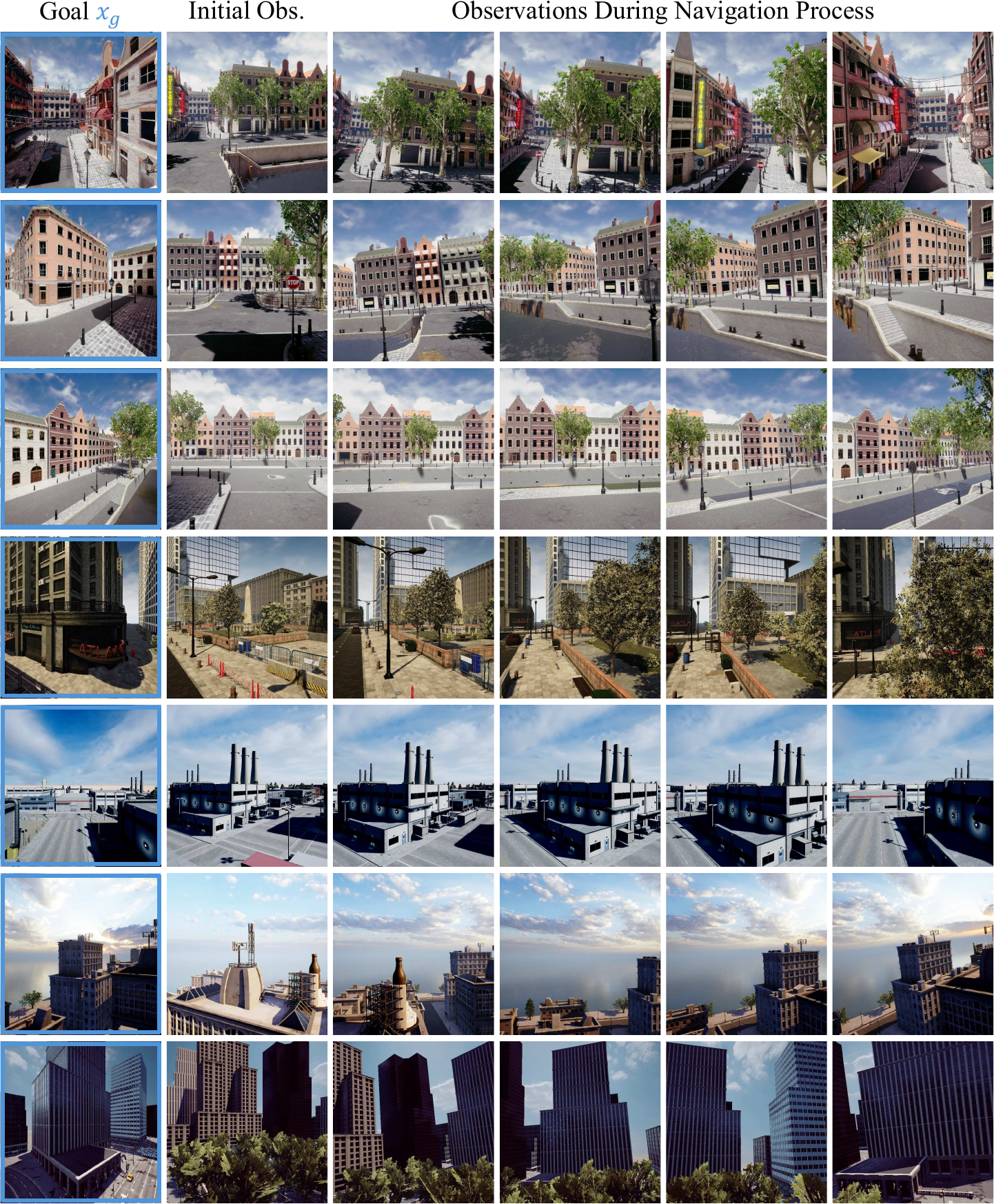}
\caption{Qualitative results of online closed-loop navigation experiments.}
\label{visual_online}
\end{figure*}

Qualitative results from the online simulation experiments are shown in Figure~\ref{visual_online}. Given a goal image with an unknown target distance, UA-NWM performs multi-step CEM planning and progressively approaches the target location through closed-loop planning.

\subsection{Real-world UAV Navigation}
\begin{figure*}[t]
\centering
\includegraphics[width=0.9\textwidth]{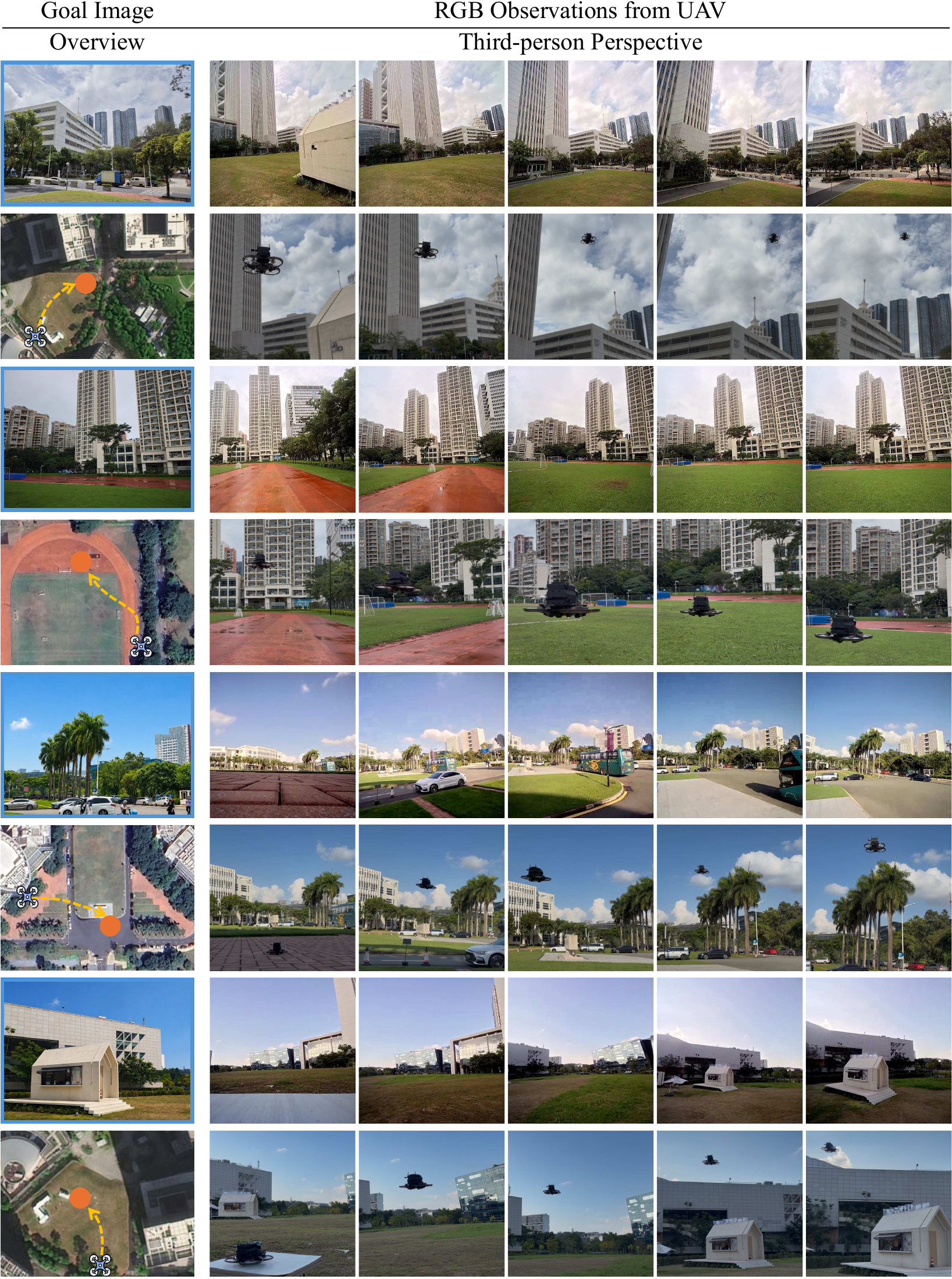}
\caption{Additional qualitative results of real-world UAV experiments.}
\label{visual_real}
\end{figure*}

Additional qualitative results from the real-world UAV experiments are shown in Figure~\ref{visual_real}. Across varying weather conditions, including sunny, cloudy, and post-rain conditions, and different times of day, including morning, afternoon, and dusk, UA-NWM demonstrates strong sim-to-real transfer and robust generalization in real-world scenes.

